\pdfoutput=1
\documentclass[11pt]{article}

\usepackage[preprint]{acl}

\usepackage{times}
\usepackage{latexsym}

\usepackage[T1]{fontenc}
\usepackage[utf8]{inputenc}

\usepackage{microtype}

\usepackage{inconsolata}

\usepackage{graphicx}

\usepackage{booktabs}
\usepackage{multirow}
\usepackage{colortbl}  %彩色表格需要加载的宏包
\usepackage{xcolor}
\usepackage{array}   %对表列和表格线的设置需要用到array宏包
\usepackage{enumerate}
\usepackage{epstopdf}
\usepackage{amsfonts,amssymb,amsmath}
\usepackage{subfigure}
\usepackage{bbm}
\usepackage{pifont}
\usepackage{listings}% 引入listings包，用于在文档中插入代码，并可自定义代码样式
\usepackage{color}
\usepackage{wrapfig}
\usepackage{url}

\definecolor{dkgreen}{rgb}{0,0.6,0}
\definecolor{gray}{rgb}{0.5,0.5,0.5}
\definecolor{mauve}{rgb}{0.58,0,0.82}
\definecolor{lemon}{RGB}{254,255,174}
\definecolor{lemongreen}{RGB}{223,247,155}
\definecolor{lightblue}{RGB}{223,247,255}
\definecolor{lightgreen}{RGB}{241,255,200}
\definecolor{lightgray}{RGB}{227,227,227}
\definecolor{lightlightgray}{RGB}{240,240,240}
\definecolor{lightorange}{RGB}{255,237,224}
\definecolor{lightpink}{RGB}{255,235,246}
\definecolor{midgray}{RGB}{0,0,0}
\definecolor{midgray2}{RGB}{43,178,29}

\newcommand{\modi}{\textcolor{black}}

\title{Dual-Space Knowledge Distillation for Large Language Models}

\author{Songming Zhang, Xue Zhang, Zengkui Sun, \textbf{Yufeng Chen}\thanks{ \ \ Yufeng Chen is the corresponding author.}, and 
\textbf{Jinan Xu}\\
Beijing Key Lab of Traffic Data Analysis and Mining, \\
Beijing Jiaotong University, Beijing, China \\
\texttt{\{smzhang22,zhang\_xue,zengksun,chenyf,jaxu\}@bjtu.edu.cn}}

\begin{document}
\maketitle
\begin{abstract}
Knowledge distillation (KD) is known as a promising solution to compress large language models (LLMs) via transferring their knowledge to smaller models.
During this process, white-box KD methods usually minimize the distance between the output distributions of the two models so that more knowledge can be transferred.
However, in the current white-box KD framework, the output distributions are from the respective output spaces of the two models, using their own prediction heads.
We argue that the space discrepancy will lead to low similarity between the teacher model and the student model on both representation and distribution levels.
Furthermore, this discrepancy also hinders the KD process between models with different vocabularies, which is common for current LLMs. 
To address these issues, we propose a dual-space knowledge distillation (DSKD) framework that unifies the output spaces of the two models for KD.
On the basis of DSKD, we further develop a cross-model attention mechanism, which can automatically align the representations of the two models with different vocabularies.
Thus, our framework is not only compatible with various distance functions for KD (\emph{e.g.}, KL divergence) like the current framework, but also supports KD between any two LLMs regardless of their vocabularies.
Experiments on task-agnostic instruction-following benchmarks show that DSKD significantly outperforms the current white-box KD framework with various distance functions, and also surpasses existing KD methods for LLMs with different vocabularies\footnote{Our code is publicly available at \url{https://github.com/songmzhang/DSKD}.}.

\end{abstract}

\section{Introduction}
Existing large language models (LLMs) have exhibited strong generalization abilities on various tasks due to their huge model capacities \cite{chowdhery23palm,touvron23llama,openai23gpt4}. 
With faith in the scaling law \cite{kaplan20scaling}, the amount of parameters in current LLMs is expanded steadily to achieve higher intelligence.
However, the increasing parameters also bring high deployment costs in real scenarios.
For this problem, knowledge distillation (KD) \cite{hinton15kd} is one of the promising solutions to compress large models with acceptable performance sacrifice.
During the process of KD, the large model typically serves as the teacher and provides supervision signals for a small model (known as the student), and thus the knowledge and the abilities of the teacher can be transferred to the lightweight student. 

Currently, KD algorithms for LLMs are usually under two frameworks, \emph{i.e.}, black-box KD and white-box KD.
Black-box KD uses the teacher's decoding sequences as the training data of the student and \modi{directly optimizes the cross-entropy loss on the one-hot target}. \cite{kim16seqkd,fu23cotkd,li23symbolickd}.
By contrast, white-box KD methods usually minimize the distance (\emph{e.g.}, KL divergence) between the output distributions of the teacher and the student, which theoretically transfer more information and usually perform better than black-box KD \cite{wen23fdiv,gu23minillm,ko24distillm}.
Although the framework of white-box KD has shown its superiority, the distributions of the student and the teacher in this framework are from different output spaces since they are produced by different prediction heads. 
At the beginning of this work, we first reveal two inherent limitations in this framework due to the discrepancy of output spaces:
\begin{itemize}
    \item \textbf{Low Teacher-Student Similarity:} The current framework usually yields low similarity between the teacher and the student on both representation and distribution levels (\S\ref{sec:low_sim}); 
    \item \textbf{Requirements on the Same Vocabulary:} A key condition for current white-box KD is that the two models should share the same vocabulary, which, however, is hardly satisfied for various LLMs in this era (\S\ref{sec:depend_same_vocab}).
\end{itemize}

% 到底是什么space，representation space or embedding space
Towards these limitations, we then propose a new framework for white-box KD, named dual-space knowledge distillation (DSKD), which is as simple as the current white-box KD framework but addresses the issues due to the space discrepancy.
Specifically, DSKD unifies the output spaces of the two models by projecting the output hidden states\footnote{In this paper, ``output hidden states'' means the hidden states output by the last layer of the model.} of the teacher/student to the representation spaces of the student/teacher, where we can use the shared prediction heads to produce the two distributions in the same output spaces.
In particular, for models with different vocabularies, we further develop a cross-model attention (CMA) mechanism to automatically align the tokens in two differently tokenized sequences.
Like the current framework, DSKD is also compatible with existing distance functions for distributions, including KL divergence, JS divergence, and so on.
Meanwhile, with CMA, we can transform distributions of the two LLMs into the same shape, which makes our framework more general and can be applied to any two LLMs regardless of their vocabularies.
% Moreover, as our framework only focuses on unifying the representation space of the two models and has no dependency on certain objectives, it is naturally compatible with all the current objectives for token-level KD, such as KL divergence, reverse KL divergence and the more advanced ones \cite{ko24distillm,wu2024rethinking}.

We evaluate our framework on instruction-following benchmarks under both settings that the two LLMs have the same/different vocabularies. 
Experimental results showcase that for LLMs with the same vocabulary, our DSKD framework significantly outperforms the current white-box KD framework on various distance functions.
Moreover, DSKD with CMA surpasses all existing KD methods for LLMs with different vocabularies.

To sum up, the contributions are as follows:
\begin{itemize}
    \item We empirically reveal that the current white-box KD framework limits the similarity between the student and the teacher \modi{due to their different output spaces}. 
    \item As a solution, we propose a new framework for white-box KD, named dual-space knowledge distillation (DSKD), which \modi{unifies the output spaces of the distributions from the teacher and the student for more effective KD.}
    \item Based on DSKD, we further develop a cross-model attention mechanism to support KD between LLMs with different vocabularies.
    \item Experiments show that our DSKD framework significantly outperforms the current white-box KD framework on various distance functions and surpasses existing KD methods for LLMs with different vocabularies.
\end{itemize}

\section{Background and Preliminary Study}
\subsection{Current Framework for White-Box KD} \label{sec:background}
Given a sequence $\mathbf{x}$, current LLMs generally learn the casual language modeling objective at each token position $i$ via the cross-entropy loss:
\begin{equation} \label{eq:ce_loss}
    \mathcal{L}_{ce} = - \sum_{i}^{|\mathbf{x}|} \log q_{\theta}(x^{*}_i|\mathbf{x}_{<i}), 
\end{equation}
where $q_{\theta}(x^{*}_i|\mathbf{x}_{<i})$ denotes the probability of the student model on the target token $x^{*}_i$ conditioning on the context $\mathbf{x}_{<i}$.
On this basis, the current white-box KD framework first feeds this sequence into the teacher model to obtain its token-level probability distributions $p(x_i|\mathbf{x}_{<i})$.
Then, the following loss is minimized to push the student distribution $q_{\theta}(x_i|\mathbf{x}_{<i})$ to the teacher distribution $p(x_i|\mathbf{x}_{<i})$:
\begin{equation} \label{eq:kd_loss}
    \mathcal{L}_{kd}=\sum_i \mathcal{D}(p(x_i|\mathbf{x}_{<i};\tau) || q_{\theta}(x_i|\mathbf{x}_{<i};\tau)),
\end{equation}
where $\mathcal{D}(\cdot || \cdot)$ is the distance function that measures the distance between the two distributions (\emph{e.g.}, KL divergence) and $\tau$ is the temperature coefficient to control the sharpness of the distributions.
% The final loss is usually the interpolation of the two losses controlled by a hyper-parameter $\alpha$:
% \begin{equation} \label{eq:ce_kd_loss}
%     \mathcal{L} = (1 - \alpha) \mathcal{L}_{ce} + \alpha \mathcal{L}_{kd}.
% \end{equation}

On the choice of the distance function $\mathcal{D}(\cdot || \cdot)$ in Eqn. \eqref{eq:kd_loss}, there have been several explorations (\emph{e.g.}, reverse KL divergence) in recent literature that aim to improve the performance of KD for LLMs \cite{wen23fdiv,agarwal24gkd,ko24distillm,wu2024rethinking}.
% Within this white-box KD framework, all these explorations focus on the formulations of the distance calculation given the two probability distributions like Eqn. \ref{eq:kd_loss}.
However, in the following section, we will uncover that no matter which distance function is employed, the current white-box KD framework has two inherent limitations since the two distributions $p(x_i|\mathbf{x}_{<i};\tau)$ and $q_{\theta}(x_i|\mathbf{x}_{<i};\tau)$ are from different output spaces.

\subsection{Limitations of the Current Framework}
\subsubsection{Low Teacher-Student Similarity} \label{sec:low_sim}
In the current white-box KD framework, the two output distributions in Eqn. \eqref{eq:kd_loss} are calculated from different output spaces of two models using their respective prediction heads.
Then, the student distribution will be optimized toward the teacher distribution by minimizing their distance.
However, we suspect this practice will limit the final similarity between the student and the teacher from two aspects: \textbf{a) representation:} as the distributions are the results of the output hidden states through the prediction heads, if the prediction heads of the two models are different, even if the distributions are close, their hidden states will not be similar; \textbf{b) distribution:} If the output hidden states of the student and the teacher are not similar, the practical distance between their distributions is difficult to reach its theoretical minimum during optimization.

\begin{figure}
	\centering
	\subfigure[Before KD]{
		\begin{minipage}[t]{0.47\linewidth}
			\centering
			\includegraphics[width=\linewidth]{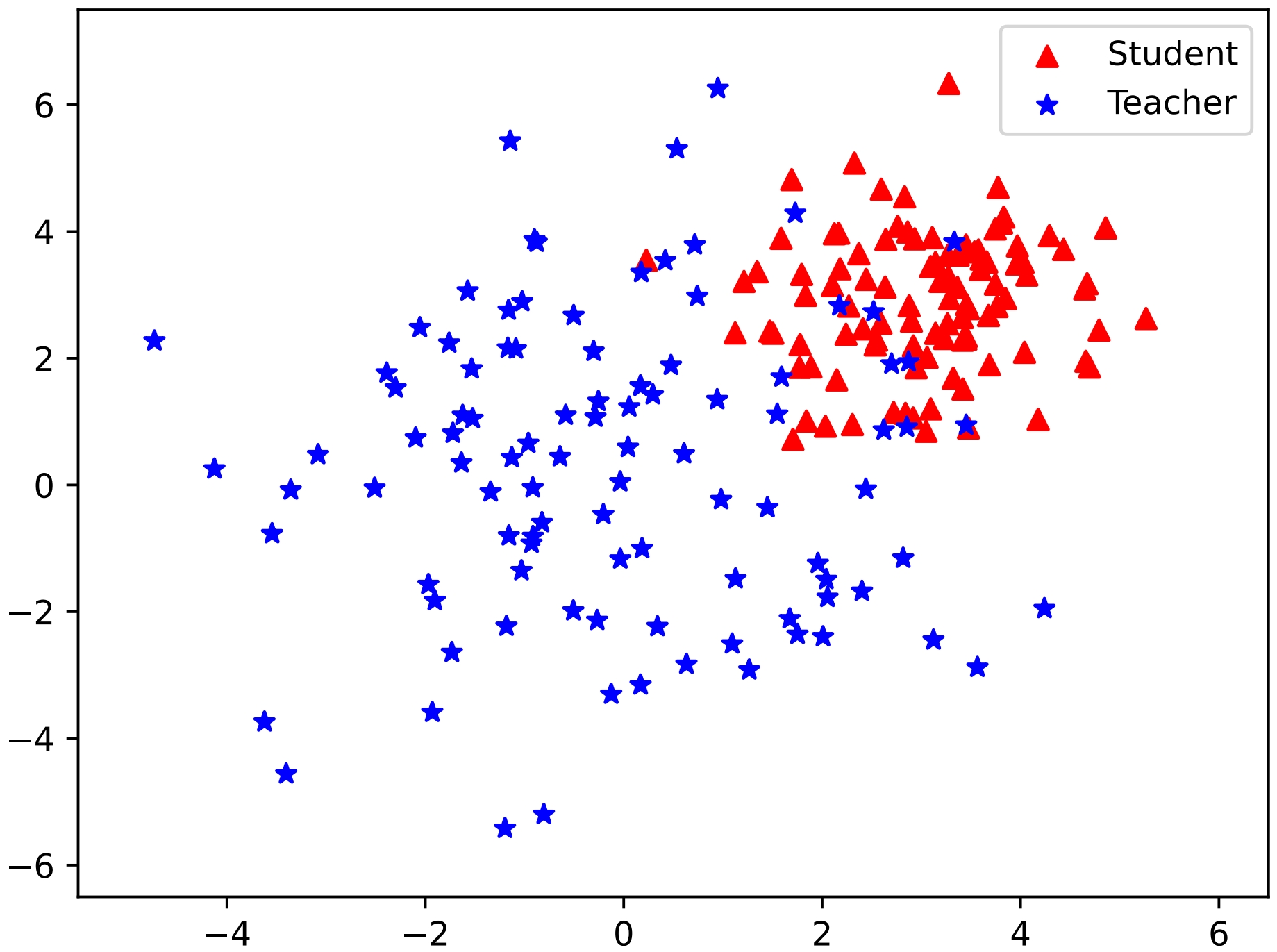}
		\end{minipage}
	}%
	\subfigure[After KD (different heads)]{
		\begin{minipage}[t]{0.47\linewidth}
			\centering
			\includegraphics[width=\linewidth]{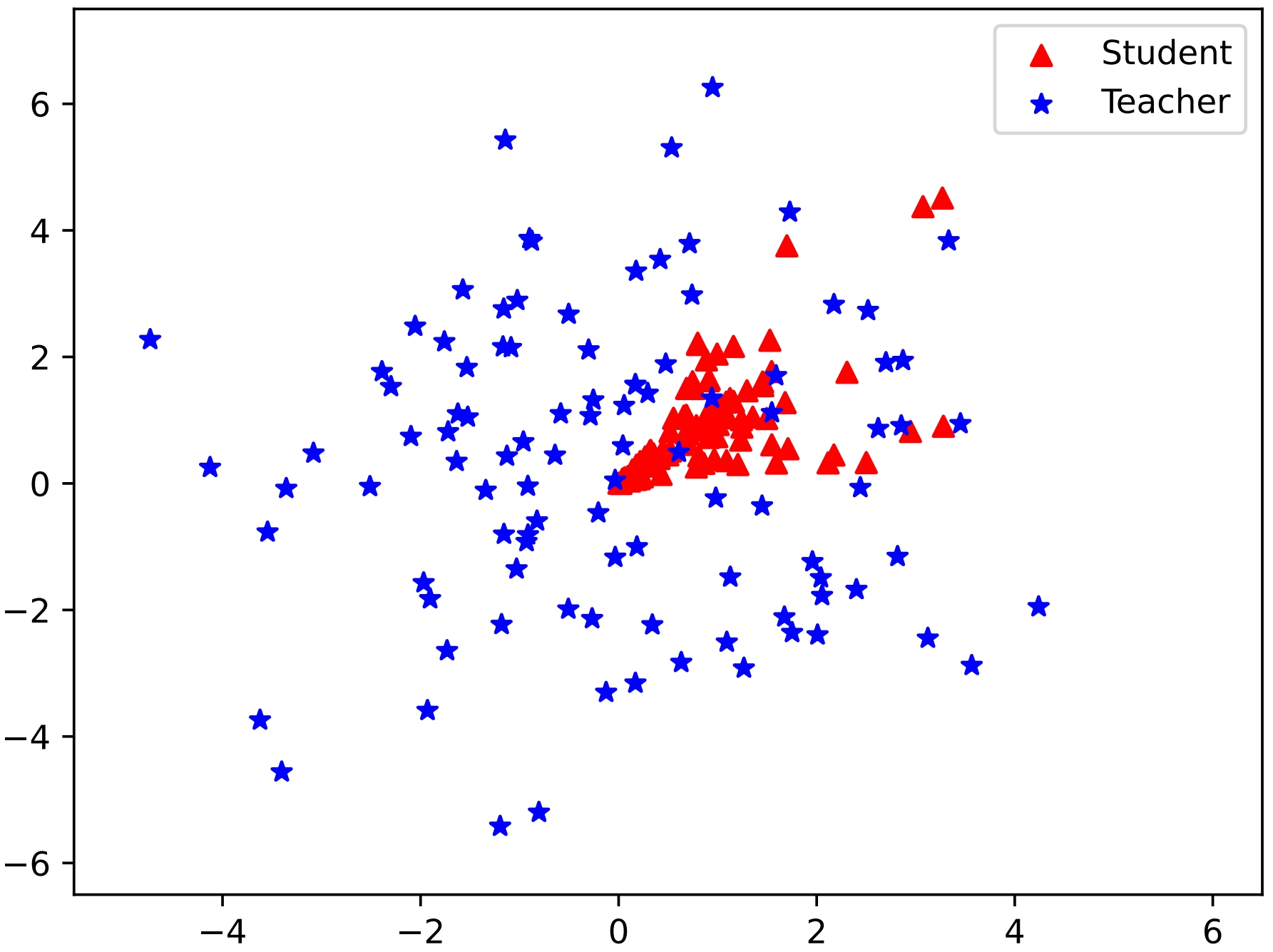}
		\end{minipage}
	}%
	%此处的空行很重要，想让图片在什么地方换行就在代码对应位置空行
 
	\subfigure[After KD (shared head)]{
		\begin{minipage}[t]{0.47\linewidth}
			\centering
			\includegraphics[width=\linewidth]{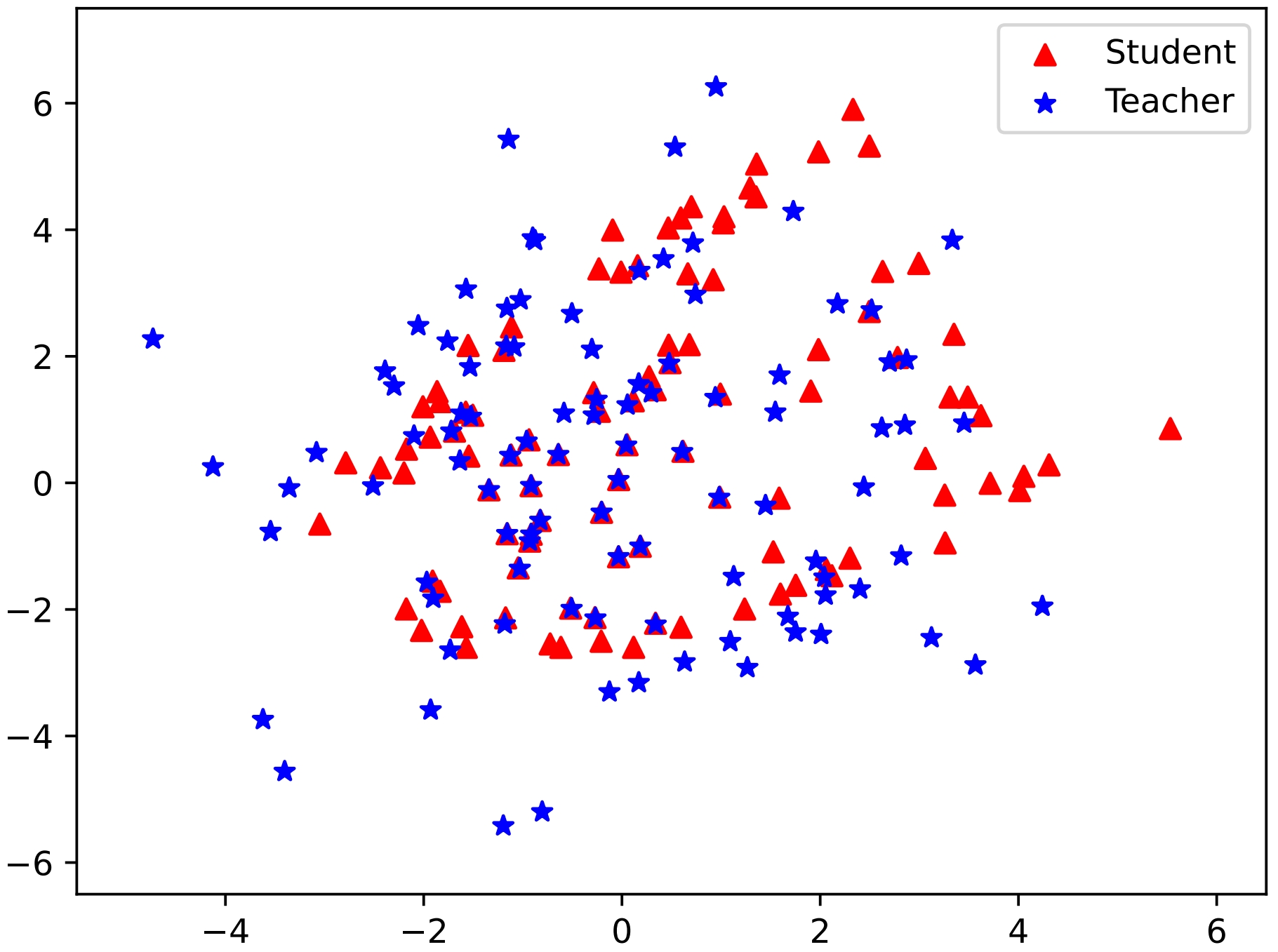}
		\end{minipage}
	}%
	\subfigure[Loss curves of KD]{
		\begin{minipage}[t]{0.47\linewidth}
			\centering
			\includegraphics[width=\linewidth]{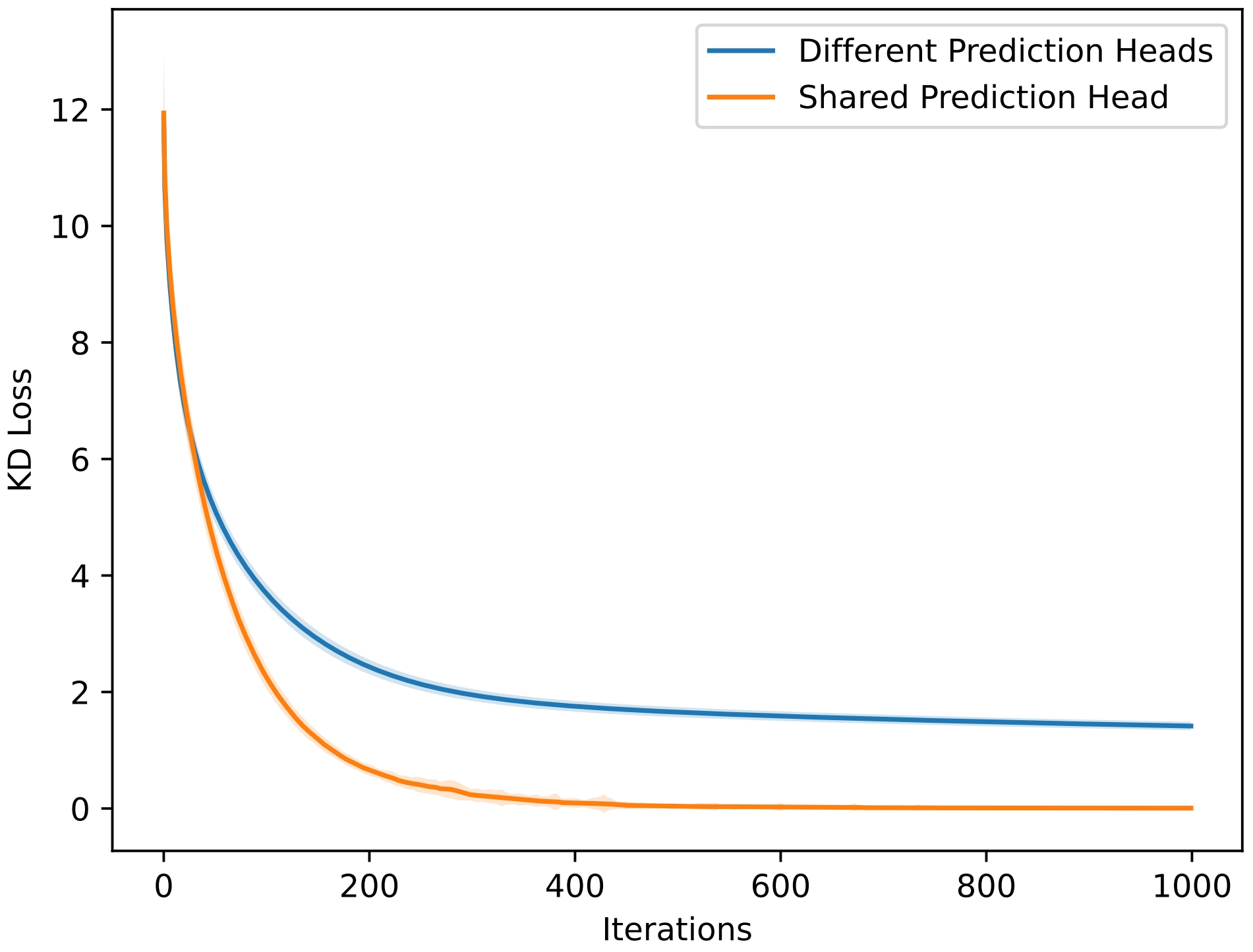}
		\end{minipage}
	}%
	\centering
	\caption{Simulation results with KL divergence as the distance function $\mathcal{D}(\cdot||\cdot)$. (a), (b) and (c) plot the {\color{red} student's hidden states} and the {\color{blue} teacher's hidden states} before and after the two KD processes. (d) shows the convergence curves of $\mathcal{L}_{kd}$ in the two KD processes.}
	\label{fig:kl_simulation}
\end{figure}

We verify the above conjectures by a simulation experiment.
In this experiment, we randomly initialize two sets of 2-D vectors (one is trainable and the other is frozen) \modi{with different mean values and variances} to represent the output hidden states of the student and the teacher, respectively (as plotted in Figure \ref{fig:kl_simulation}(a)).
Besides, we set two prediction heads to produce probability distributions of the student and the teacher from these vectors.
Then, we select KL divergence as the distance function $\mathcal{D}(\cdot||\cdot)$ and simulate the KD process with $\mathcal{L}_{kd}$ in Eqn. \eqref{eq:kd_loss} for 1000 iterations.
After the iterations, we plot the two sets of vectors again and record the loss curve during the whole process in Figure \ref{fig:kl_simulation}.

Firstly, we simulate the process of the current white-box KD framework, which uses distributions from different output spaces produced by different prediction heads.
The result in Figure \ref{fig:kl_simulation}(b) shows that the student's hidden states optimized by the current KD framework exhibit distinct structure discrepancy from the teacher's hidden states, reflecting low similarity between them.
As a comparison, we then unify the output spaces of the two distributions by sharing the same prediction head for the student and the teacher and conduct the same KD process as above.
As shown in Figure \ref{fig:kl_simulation}(c), under this setting, the student's hidden states become more similar and closer to the teacher's hidden states.
The significant difference between these two settings indicates that the current KD framework may lead to sub-optimal similarity between the student and the teacher \textbf{on the representation level}.
By contrast, a better alternative is to unify the output spaces for the distributions of the student and the teacher.

Then, we repeat the simulations of the above two settings 100 times and plot their averaged curves of $\mathcal{L}_{kd}$ in Figure \ref{fig:kl_simulation}(d).
As we suspected, when using different prediction heads, the value of KL divergence still leaves a large margin to its theoretical minimum (\emph{i.e.}, 0) after convergence.
On the contrary, when using a shared prediction head, the value of KL divergence will converge faster and finally be closer to this minimum.
It sufficiently illustrates that the current KD framework also limits the similarity between the two models \textbf{on the distribution level}.
Besides KL divergence, we also conduct these simulations with other distance functions (\emph{e.g.}, reverse KL divergence, JS divergence, etc.).
The results are shown in Appendix \ref{sec:other_simulation}\modi{, which} also support the above conclusions.
Additionally, we provide the pseudo code of the simulation experiment in Appendix \ref{sec:pseudo_code} to present more details.

\subsubsection{Dependency on the Same Vocabulary} \label{sec:depend_same_vocab}
As stated in \S\ref{sec:background}, the current KD framework minimizes the distance between the two distributions at each token position.
However, when the teacher and the student have different vocabularies, the same text may be tokenized into different sequences like $\mathbf{x}=[x_1,x_2,...,x_n]$ and $\mathbf{y}=[y_1,y_2,...,y_m]$.
Under this circumstance, the teacher distribution $p(y_i|\mathbf{y}_{<i})$ is probably incorrect for $q_{\theta}(x_i|\mathbf{x}_{<i})$.
Additionally, as the output spaces are more different when the prediction heads contain different vocabularies, the produced distributions are even with different dimensions, which is obviously prohibited by Eqn. \eqref{eq:kd_loss}.
Therefore, the current white-box KD framework fails to work between LLMs with different vocabularies.
% \begin{figure}
%     \centering
%     \includegraphics[width=0.5\linewidth]{before.eps}
%     \caption{Before KD}
%     \label{fig:enter-label}
% \end{figure}
% \begin{figure}
%     \centering
%     \includegraphics[width=\linewidth]{after_kl.eps}
%     \caption{After KL from different spaces}
%     \label{fig:enter-label}
% \end{figure}
% \begin{figure}
%     \centering
%     \includegraphics[width=\linewidth]{after_stu_kl.eps}
%     \caption{After KL from the same space}
%     \label{fig:enter-label}
% \end{figure}

\section{Methodology}
This section introduces our solutions to the above limitations of the current white-box KD framework.
Firstly, we will introduce our new KD framework in \S\ref{sec:method_framework}.
Then we present a cross-model attention mechanism in \S\ref{sec:method_cross_model} to extend our framework to support LLMs with different vocabularies.
\subsection{Dual-Space Knowledge Distillation Framework} 
\label{sec:method_framework}
Inspired by the observations in \S\ref{sec:low_sim}, we design our dual-space knowledge distillation (DSKD) framework.
The core idea is to unify the output spaces of the two distributions in Eqn. \eqref{eq:kd_loss}.
To achieve this, we project the output hidden states of the teacher/student model into the representation space of the student/teacher model, so that the distributions can be output by the same prediction head and thus lie in \textbf{the unified output space}.
Next, we will detail how to conduct the projection and unify KD in student and teacher space.

\paragraph{KD in Student Space.} In the student space, we first use a linear projector $\mathcal{P}^{t \rightarrow s}$ to transform the hidden states of the teacher model into the representation space of the student model.
Here, we denote the output hidden states of the whole sequence from the teacher model as $\mathbf{h}_{1:n}^{t}$.
Then the projection process can be formulated as follows:
\begin{equation} \label{eq:down_transform}
    \mathbf{h}_{1:n}^{t \rightarrow s} = \mathcal{P}^{t \rightarrow s}(\mathbf{h}_{1:n}^{t}; \theta_{\mathcal{P}}^{t \rightarrow s}) \in \mathbb{R}^{n \times d},
\end{equation}
where $\theta_{\mathcal{P}}^{t \rightarrow s}$ is the trainable parameter of the projector $\mathcal{P}^{t \rightarrow s}$ and $d$ is the hidden size of the student model.
With the projected hidden states $\mathbf{h}^{t \rightarrow s}$, we can obtain the transformed teacher distribution $\mathbf{p}_{1:n}^{t \rightarrow s}$ that shares the same output space with the student using the student's prediction head $\mathbf{W}^s \in 
\mathbb{R}^{d \times |V|}$:
\begin{equation} \label{eq:p_t2s}
    \mathbf{p}_{1:n}^{t \rightarrow s} = {\rm softmax}(\mathbf{h}_{1:n}^{t \rightarrow s}\mathbf{W}^s) \in \mathbb{R}_{+}^{n \times |V|},
\end{equation}
where $|V|$ is the vocabulary size of the two models.
As the projector is randomly initialized at the start of the training, we train the transformed distribution $\mathbf{p}_{1:n}^{t \rightarrow s}$ to predict the ground-truth target tokens in the student's sequence with cross-entropy loss\footnote{Note that we stop the gradient of $\mathbf{W}^s$ in Eqn. \eqref{eq:p_t2s} to avoid negative effects to the student model}:
\begin{equation} \label{eq:stuside_ce_loss}
    \mathcal{L}^{t \rightarrow s}_{ce}=- \sum_i \log (p^{t \rightarrow s}(x^{*}_i|\mathbf{x}_{<i})).
\end{equation}
Meanwhile, we use this distribution $p^{t \rightarrow s}$ as the new teacher distribution and calculate the same loss for KD as Eqn. \eqref{eq:kd_loss}:
\begin{equation} \label{eq:stuside_kd_loss}
    \mathcal{L}^{stu}_{kd}=\sum_i \mathcal{D}(p^{t \rightarrow s}(x_i|\mathbf{x}_{<i};\tau) || q_{\theta}(x_i|\mathbf{x}_{<i};\tau)),
\end{equation}
where $\mathcal{D}(\cdot||\cdot)$ is as same as the one in Eqn. \eqref{eq:kd_loss}. Note that we stop the gradient of $p^{t \rightarrow s}(x_i|\mathbf{x}_{<i};\tau)$ in Eqn. \eqref{eq:stuside_kd_loss} so that $\mathcal{L}^{stu}_{kd}$ will not collapse.

\paragraph{KD in Teacher Space.} Similar to the process in the student space, we also project the hidden states of the student model into the teacher's dimension using another projector $\mathcal{P}^{s \rightarrow t}$:
\begin{equation}
    \mathbf{h}_{1:n}^{s \rightarrow t} = \mathcal{P}^{s \rightarrow t}(\mathbf{h}_{1:n}^{s}; \theta_{\mathcal{P}}^{s \rightarrow t}) \in \mathbb{R}^{n \times D},
\end{equation}
where $D$ is the hidden size of the teacher model.
Then, we use the prediction head of the teacher model $\mathbf{W}^t \in \mathbb{R}^{D \times |V|}$ to obtain the distributions of the student model in the teacher's space:
\begin{equation} \label{eq:p_s2t}
    \mathbf{q}_{\theta}{}_{1:n}^{s \rightarrow t} = {\rm softmax}(\mathbf{h}_{1:n}^{s \rightarrow t}\mathbf{W}^t) \in \mathbb{R}_{+}^{n \times |V|},
\end{equation}
As the teacher distributions in its own space are usually well-trained, we can directly calculate the KD loss in the teacher space:
\begin{equation} \label{eq:teaside_kd_loss}
    \mathcal{L}^{tea}_{kd}=\sum_i {\rm KL}(p(x_i|\mathbf{x}_{<i};\tau) || q_{\theta}^{s \rightarrow t}(x_i|\mathbf{x}_{<i};\tau)),
\end{equation}
where a difference from Eqn. \eqref{eq:stuside_kd_loss} is that we directly fix KL divergence as $\mathcal{D}(\cdot||\cdot)$ since we found it more appropriate for KD in the teacher space.

The whole loss of DSKD sums the KD losses in both spaces and the cross-entropy loss in Eqn. \eqref{eq:stuside_ce_loss}:
\begin{equation} \label{eq:dskd_loss}
    \mathcal{L}_{dskd}=\mathcal{L}^{stu}_{kd} + \mathcal{L}^{tea}_{kd} + \mathcal{L}^{t \rightarrow s}_{ce}.
\end{equation}

\subsection{Cross-Model Attention Mechanism}
\label{sec:method_cross_model}
% \subsubsection{Cross-Model Attention}
In the above section, we have introduced our DSKD framework for LLMs with the same vocabulary.
For LLMs with different vocabularies, since DSKD always produces distributions with the same dimensions for the student and the teacher via sharing the same prediction heads, the remaining requirement for KD is just to align the tokens in the two sequences tokenized by different tokenizers\footnote{Here we borrow the notations in \S\ref{sec:depend_same_vocab} and assume that there are $m$ tokens in the teacher's sequence.}.

To this end, we develop a cross-model attention (CMA) mechanism to learn the alignment between tokens in the two sequences automatically.
Specifically, we first concatenate the \modi{student's embeddings of input tokens $\mathbf{e}_{1:n}^s$ and target tokens $\mathbf{e}_{2:n+1}^s$ in the sequence on the last dimension and project them} as the query vectors with a query projector $\mathcal{P}^q$:
\begin{equation}
Q=\mathcal{P}^q([\mathbf{e}_{1:n}^s;\mathbf{e}_{2:n+1}^s]; \theta^{q}_{\mathcal{P}}) \in \mathbb{R}^{n\times 2D}. \nonumber
\end{equation}
Similarly, we use the teacher's embeddings and output hidden states to obtain the key and value vectors:
\begin{align*}
    K&={\rm N}([\mathbf{e}_{1:m}^t;\mathbf{e}_{2:m+1}^t])  \in \mathbb{R}^{m \times 2D}, \\
    V&=\mathcal{P}^{v}({\rm  N}(\mathbf{e}_{2:m+1}^t)+{\rm N}(\mathbf{h}_{1:m}^t); \theta^{v}_{\mathcal{P}}) \in \mathbb{R}^{m \times d},
\end{align*}
where we normalize the embeddings and the hidden states of the teacher with their standard deviations like ${\rm N}(x)=x/{\rm std}(x)$ for faster convergence.
% \begin{align*}
%     \mathbf{\tilde{e}}^t&=\mathbf{e}^t / \sigma(\mathbf{e}^t), \\
%     \mathbf{\tilde{h}}^t&=\mathbf{h}^t / \sigma(\mathbf{h}^t).
% \end{align*}
% \begin{equation}
%     \mathbf{\tilde{e}}^t=\mathbf{e}^t / \sigma(\mathbf{e}^t), \quad\mathbf{\tilde{h}}^t=\mathbf{h}^t / \sigma(\mathbf{h}^t). \nonumber
% \end{equation}

Then, we calculate the attention matrix with the query and the key:
\begin{equation}
    \mathbf{a}^{t \rightarrow s}={\rm softmax}(\frac{QK^{\top}}{\sqrt{2D}}) \in \mathbb{R}^{n \times m}.
\end{equation}
The attention matrix reflects the alignment relationship from the teacher tokens to the student tokens.
Based on this matrix, we can obtain the final projected and aligned hidden states of the teacher model from the weighted sum of the value vectors:
\begin{equation}
    \mathbf{\tilde{h}}_{1:n}^{t \rightarrow s}=\mathbf{a}^{t \rightarrow s}V \in \mathbb{R}^{n \times d}.
\end{equation}
Then, we can substitute $\mathbf{\tilde{h}}^{t \rightarrow s}$ into Eqn. \eqref{eq:p_t2s} and train $\mathbf{\tilde{h}}^{t \rightarrow s}$ to correctly predict the target tokens of the student model with Eqn. \eqref{eq:stuside_ce_loss}.
Meanwhile, the teacher distributions produced from $\mathbf{\tilde{h}}^{t \rightarrow s}$ are also in the student space and can support the KD process in Eqn. \eqref{eq:stuside_kd_loss}\footnote{For models with different vocabularies, the distribution in Eqn. \eqref{eq:p_t2s} usually has lower accuracy, so we mask the KD loss in Eqn. \eqref{eq:stuside_kd_loss} when the teacher distribution is incorrect.}.

Besides, we also transpose the matrix to align the student tokens to the teacher tokens:
\begin{equation}
    \mathbf{a}^{s \rightarrow t}={\rm softmax}(\frac{KQ^{\top}}{\sqrt{2D}}) \in \mathbb{R}^{m \times n}.
\end{equation}
We can project and align the student's hidden states to the teacher's using this alignment matrix:
\begin{equation}
    \mathbf{\tilde{h}}_{1:m}^{s \rightarrow t}=\mathbf{a}^{s \rightarrow t}\mathcal{P}^{s \rightarrow t}(\mathbf{h}_{1:n}^s;\theta^{s \rightarrow t}_{\mathcal{P}}) \in \mathbb{R}^{m \times D}.
\end{equation}
Then, we can substitute $\mathbf{\tilde{h}}_{1:m}^{s \rightarrow t}$ into Eqn. \eqref{eq:p_s2t} and conduct KD in the teacher space with Eqn. \eqref{eq:teaside_kd_loss}.

% \subsubsection{Gaussian Position Bias}
% To prevent the query from attending the key in the wrong position, we introduce a Gaussian position bias.

\section{Experiments}
\subsection{Experimental Setup}
\paragraph{Data.} We evaluate our DSKD framework on several instruction-following datasets following \citet{gu23minillm}.
Specifically, we choose $\mathtt{databricks}$-$\mathtt{dolly}$-$\mathtt{15k}$ dataset processed by \citet{gu23minillm} to conduct the KD process, which contains about 11k samples for training, 1k for validation, and 500 for testing.
Besides, we also select Self-Instruct (\textbf{SelfInst}), Vicuna-Evaluation ({\bf VicunaEval}), Super Natural Instructions ({\bf S-NI}), and Unnatural Instructions ({\bf UnNI}) as the additional test sets for more comprehensive evaluation.
\paragraph{Models.} For student LLMs, we select both GPT2-120M \cite{radford19gpt2} and TinyLLaMA-1.1B \cite{zhang24tinyllama}.
For GPT2-120M, we employ GPT2-1.5B and Qwen1.5-1.8B \cite{bai23qwen} respectively as the teacher LLMs that have the same/different vocabularies with the student LLMs.
For TinyLLaMA-1.1B, we choose LLaMA2-7B \cite{touvron23llama} and Mistral-7B \cite{jiang23mistral} as the teacher LLMs that have the same/different vocabularies with the student LLMs.
\paragraph{Training and Evaluation.} For KD on GPT2, we employ full-finetuning for the teachers and the students.
For KD on TinyLLaMA, we finetune the students and the teachers with LoRA.
In particular, we set the temperature $\tau$ to 2.0 according the performance on the validation set.
Besides, all the projectors in our method are linear layers, which only increase few parameters in training (\emph{e.g.}, $\approx$2M for DSKD on GPT2). 
For the evaluation, we sampling the responses from the models under 5 random seeds.
The final performance is measured by Rouge-L \cite{lin04rouge} between the generated responses and the human-labeled ones. More details are provided in Appendix \ref{sec:experiment_detail}.

\begin{table*}[t]
    \centering
    \resizebox{0.96\linewidth}{!}{
        \begin{tabular}{lccccc|c}
            \bottomrule
            \textbf{Methods} & \textbf{Dolly} & \textbf{SelfInst} & \textbf{VicunaEval} & \textbf{S-NI} & \textbf{UnNI} & \textbf{Avg.} \\
            \hline
            \hline
            SFT & 22.94$_{\pm 0.28}$ & 10.11$_{\pm 0.36}$ & 15.17$_{\pm 0.63}$ & 16.21$_{\pm 0.19}$ & 18.68$_{\pm 0.09}$ & 16.62 \\
            \hline
            % \rowcolor{lightblue}
            % \rowcolor{lightgray}
            \multicolumn{7}{c}{\textbf{GPT2-1.5B $\rightarrow$ GPT2-120M (Same Vocabulary)}} \\
            \hline
            \rowcolor{lightgray}
            {\color{midgray} Teacher} & {\color{midgray} 27.19$_{\pm 0.23}$} & {\color{midgray} 14.64$_{\pm 0.64}$} & {\color{midgray} 16.30$_{\pm 0.37}$} & {\color{midgray} 27.55$_{\pm 0.30}$} & {\color{midgray} 31.42$_{\pm 0.11}$} & {\color{midgray} 23.42} \\
            \hline
            SeqKD & 23.68$_{\pm 0.25}$ & 10.03$_{\pm 0.23}$ & 14.41$_{\pm 0.46}$ & 16.36$_{\pm 0.18}$ & 18.48$_{\pm 0.11}$ & 16.59 \\
            \hline
            KL & 24.54$_{\pm 0.48}$ & 10.43$_{\pm 0.24}$ & 15.66$_{\pm 0.42}$ & 17.24$_{\pm 0.27}$ & 20.28$_{\pm 0.18}$ & 17.63 \\
            % \rowcolor{lightgreen}
            \quad \emph{w/} DSKD (ours) & 24.70$_{\pm 0.24}$ & 10.65$_{\pm 0.30}$ & 15.67$_{\pm 0.30}$ & 19.51$_{\pm 0.21}$ & 22.94$_{\pm 0.07}$ & 18.69 {\color{midgray2} \small(+1.06$\uparrow$)} \\
            \hline
            RKL & 24.38$_{\pm 0.55}$ & 10.73$_{\pm 0.61}$ & 15.71$_{\pm 0.39}$ & 17.31$_{\pm 0.11}$ & 20.96$_{\pm 0.12}$ & 17.82 \\
            % \rowcolor{lightgreen}
            \quad \emph{w/} DSKD (ours) & 24.61$_{\pm 0.59}$ & 11.01$_{\pm 0.45}$ & 14.98$_{\pm 0.48}$ & 19.32$_{\pm 0.28}$ & 22.27$_{\pm 0.13}$ & 18.44 {\color{midgray2} \small(+0.62$\uparrow$)} \\
            \hline
            JS & 23.86$_{\pm 0.14}$ & 10.20$_{\pm 0.40}$ & 15.50$_{\pm 0.23}$ & 16.20$_{\pm 0.23}$ & 19.17$_{\pm 0.06}$ & 16.98 \\
            % \rowcolor{lightgreen}
            \quad \emph{w/} DSKD (ours) & 24.61$_{\pm 0.27}$ & 11.41$_{\pm 0.35}$ & 15.40$_{\pm 0.28}$ & 18.94$_{\pm 0.20}$ & 21.48$_{\pm 0.17}$ & 18.37 {\color{midgray2} \small(+1.39$\uparrow$)} \\
            \hline
            SKL \cite{ko24distillm} & 24.03$_{\pm 0.23}$ & 10.66$_{\pm 0.51}$ & 14.70$_{\pm 0.37}$ & 17.99$_{\pm 0.15}$ & 21.18$_{\pm 0.16}$ & 17.71 \\
            % \rowcolor{lightgreen}
            \quad \emph{w/} DSKD (ours) & 25.24$_{\pm 0.28}$ & 10.50$_{\pm 0.13}$ & 15.76$_{\pm 0.43}$ & 18.34$_{\pm 0.44}$ & 20.87$_{\pm 0.11}$ & 18.14 {\color{midgray2} \small(+0.43$\uparrow$)} \\
            \hline
            SRKL \cite{ko24distillm} & 24.48$_{\pm 0.19}$ & 10.35$_{\pm 0.38}$ & 14.88$_{\pm 0.24}$ & 16.53$_{\pm 0.23}$ & 19.68$_{\pm 0.05}$ & 17.19 \\
            % \rowcolor{lightgreen}
            \quad \emph{w/} DSKD (ours) & 25.23$_{\pm 0.25}$ & 11.19$_{\pm 0.22}$ & 15.91$_{\pm 0.45}$ & 17.92$_{\pm 0.16}$ & 21.20$_{\pm 0.12}$ & 18.29 {\color{midgray2} \small(+1.10$\uparrow$)} \\
            \hline
            AKL \cite{wu2024rethinking} & 24.75$_{\pm 0.60}$ & 10.46$_{\pm 0.24}$ & 15.37$_{\pm 0.41}$ & 17.48$_{\pm 0.17}$ & 20.11$_{\pm 0.05}$ & 17.63 \\
            % DSKD (ours) & & & & & \\
            % \rowcolor{lightgreen}
            \quad \emph{w/} DSKD (ours) & 25.13$_{\pm 0.14}$ & 10.63$_{\pm 0.43}$ & 16.18$_{\pm 0.35}$ & 18.58$_{\pm 0.48}$ & 21.45$_{\pm 0.16}$ & 18.39 {\color{midgray2} \small(+0.76$\uparrow$)} \\
            \hline
            % \rowcolor{lightgray}
            % \rowcolor{lightpink} 
            \multicolumn{7}{c}{\textbf{Qwen1.5-1.8B $\rightarrow$ GPT2-120M (Different Vocabularies)}} \\
            \hline
            \rowcolor{lightgray}
            {\color{midgray} Teacher} & {\color{midgray} 27.42$_{\pm 0.33}$} & {\color{midgray} 19.42$_{\pm 0.11}$} & {\color{midgray} 19.31$_{\pm 0.21}$} & {\color{midgray} 34.87$_{\pm 0.30}$} & {\color{midgray} 36.00$_{\pm 0.10}$} & {\color{midgray} 27.40} \\
            \hline
            SeqKD & 23.40$_{\pm 0.21}$ & 9.36$_{\pm 0.38}$ & 15.37$_{\pm 0.35}$ & 15.16$_{\pm 0.17}$ & 17.34$_{\pm 0.11}$ & 16.13 \\
            MinED \cite{wan24fusellm} & 24.41$_{\pm 0.61}$ & 10.60$_{\pm 0.39}$ & 15.86$_{\pm 0.42}$ & 16.76$_{\pm 0.28}$ & 19.68$_{\pm 0.12}$ & 17.46 \\
            ULD \cite{boizard2024uld} \qquad & 23.77$_{\pm 0.41}$ & 9.67$_{\pm 0.50}$ & 14.99$_{\pm 0.55}$ & 17.60$_{\pm 0.21}$ & 19.49$_{\pm 0.12}$ & 17.11 \\
            % DSKD + Align (ours) & & & & & \\
            \hline
            DSKD-CMA-SRKL (ours) & 25.23$_{\pm 0.17}$ & 10.99$_{\pm 0.26}$ & 15.56$_{\pm 0.41}$ & 17.76$_{\pm 0.23}$ & 20.54$_{\pm 0.07}$ & 18.02 \\
            % DSKD-CMA-RKL (ours) & 23.57$_{\pm 0.14}$ & 10.87$_{\pm 0.33}$ & 14.82$_{\pm 0.35}$ & 18.67$_{\pm 0.13}$ & 20.40$_{\pm 0.11}$ & 17.66  \\
            \toprule
        \end{tabular}
    }
    \caption{Rouge-L scores (\%) on several benchmarks with GPT2-120M as the student. We list the mean values and the standard deviations among 5 random seeds. The average scores (\textbf{Avg.}) on all benchmarks are also listed. ``\emph{w/} DSKD'' denotes our DSKD using the corresponding distance function as $\mathcal{D}(\cdot||\cdot)$ in Eqn. \eqref{eq:stuside_kd_loss}. And ``DSKD-CMA-SRKL'' denotes our DSKD framework equipped with cross-model attention with SRKL as $\mathcal{D}(\cdot||\cdot)$ in Eqn. \eqref{eq:stuside_kd_loss}.}
    \label{tab:main_results_gpt2}
    \vspace{-5pt}
\end{table*}

\begin{table*}[h]
    \centering
    \resizebox{0.96\linewidth}{!}{
        \begin{tabular}{lccccc|c}
            \bottomrule
            \textbf{Methods} & \textbf{Dolly} & \textbf{SelfInst} & \textbf{VicunaEval} & \textbf{S-NI} & \textbf{UnNI} & \textbf{Avg.} \\
            \hline
            \hline
            SFT & 23.20$_{\pm 0.13}$ & 14.88$_{\pm 0.54}$ & 16.42$_{\pm 0.35}$ & 27.79$_{\pm 0.27}$ & 26.12$_{\pm 0.11}$ & 21.68 \\
            \hline
            % \rowcolor{lightblue}
            % \rowcolor{lightgray}
            \multicolumn{7}{c}{\textbf{LLaMA2-7B $\rightarrow$ TinyLLaMA-1.1B (Same Vocabulary)}} \\
            \hline
            \rowcolor{lightgray}
            Teacher & 28.32$_{\pm 0.46}$ & 20.95$_{\pm 0.69}$ & 18.76$_{\pm 0.35}$ & 32.05$_{\pm 0.28}$ & 32.41$_{\pm 0.12}$ & 26.50 \\
            \hline
            SeqKD & 23.21$_{\pm 0.22}$ & 16.46$_{\pm 0.72}$ & 16.58$_{\pm 0.38}$ & 26.33$_{\pm 0.26}$ & 27.69$_{\pm 0.10}$ & 22.05 \\
            \hline
            KL & 25.46$_{\pm 0.63}$ & 17.21$_{\pm 0.25}$ & 16.43$_{\pm 0.53}$ & 29.27$_{\pm 0.29}$ & 29.28$_{\pm 0.09}$ & 23.53 \\
            \quad \emph{w/} DSKD (ours) & 26.31$_{\pm 0.26}$ & 18.27$_{\pm 0.56}$ & 18.04$_{\pm 0.37}$ & 31.43$_{\pm 0.26}$ & 31.20$_{\pm 0.09}$ & 25.05 {\color{midgray2} \small(+1.52$\uparrow$)} \\
            \hline
            RKL & 24.49$_{\pm 0.41}$ & 17.14$_{\pm 0.61}$ & 16.87$_{\pm 0.26}$ & 29.50$_{\pm 0.28}$ & 29.36$_{\pm 0.08}$ & 23.47 \\
            \quad \emph{w/} DSKD (ours) & 26.93$_{\pm 0.34}$ & 18.14$_{\pm 0.54}$ & 18.81$_{\pm 0.39}$ & 31.79$_{\pm 0.31}$ & 32.49$_{\pm 0.11}$ & 25.63 {\color{midgray2} \small (+2.17$\uparrow$)} \\
            \hline
            JS & 24.03$_{\pm 0.31}$ & 15.75$_{\pm 0.51}$ & 16.64$_{\pm 0.30}$ & 28.08$_{\pm 0.10}$ & 28.68$_{\pm 0.08}$ & 22.62 \\
            % \rowcolor{lightgreen}
            \quad \emph{w/} DSKD (ours) & 24.79$_{\pm 0.42}$ & 17.10$_{\pm 0.47}$ & 16.78$_{\pm 0.20}$ & 29.06$_{\pm 0.18}$ & 29.47$_{\pm 0.22}$ & 23.44 {\color{midgray2} \small(+0.82$\uparrow$)} \\
            \hline
            SKL \cite{ko24distillm} & 24.14$_{\pm 0.53}$ & 15.98$_{\pm 0.72}$ & 16.89$_{\pm 0.22}$ & 29.30$_{\pm 0.18}$ & 28.71$_{\pm 0.12}$ & 23.01 \\
            \quad \emph{w/} DSKD (ours) & 25.88$_{\pm 0.22}$ & 17.59$_{\pm 0.56}$ & 17.17$_{\pm 0.34}$ & 29.52$_{\pm 0.33}$ & 30.69$_{\pm 0.16}$ & 24.17 {\color{midgray2} \small(+1.16$\uparrow$)} \\
            \hline
            SRKL \cite{ko24distillm} & 24.28$_{\pm 0.58}$ & 16.91$_{\pm 0.67}$ & 16.88$_{\pm 0.20}$ & 29.55$_{\pm 0.19}$ & 28.64$_{\pm 0.21}$ & 23.25 \\
            \quad \emph{w/} DSKD (ours) & 25.44$_{\pm 0.22}$ & 17.34$_{\pm 0.69}$ & 17.19$_{\pm 0.34}$ & 30.29$_{\pm 0.29}$ & 31.23$_{\pm 0.13}$ & 24.30 {\color{midgray2} \small(+1.05$\uparrow$)} \\
            \hline
            AKL \cite{wu2024rethinking} & 24.80$_{\pm 0.70}$ & 16.79$_{\pm 1.09}$ & 16.80$_{\pm 0.44}$ & 29.29$_{\pm 0.35}$ & 28.81$_{\pm 0.09}$ & 23.30 \\
            % DSKD (ours) & & & & & \\
            % \rowcolor{lightlightgray}
            \quad \emph{w/} DSKD (ours) & 26.33$_{\pm 0.45}$ & 20.17$_{\pm 0.46}$ & 17.43$_{\pm 0.48}$ & 34.93$_{\pm 0.39}$ & 34.40$_{\pm 0.20}$ & 26.65 {\color{midgray2} \small(+3.35$\uparrow$)} \\
            \hline
            % \rowcolor{lightpink}
            % \rowcolor{lightgray}
            \multicolumn{7}{c}{\textbf{Mistral-7B $\rightarrow$ TinyLLaMA-1.1B (Different Vocabularies)}} \\
            \hline
            \rowcolor{lightgray}
            Teacher & 31.56$_{\pm 0.19}$ & 25.10$_{\pm 0.36}$ & 20.50$_{\pm 0.32}$ & 36.07$_{\pm 0.24}$ & 36.27$_{\pm 0.15}$ & 29.90 \\
            \hline
            SeqKD & 23.56$_{\pm 0.39}$ & 15.87$_{\pm 0.54}$ & 15.99$_{\pm 0.55}$ & 25.50$_{\pm 0.37}$ & 26.64$_{\pm 0.09}$ & 21.51 \\
            MinED \cite{wan24fusellm} & 20.96$_{\pm 0.51}$ & 14.49$_{\pm 0.35}$ & 15.98$_{\pm 0.45}$ & 27.21$_{\pm 0.13}$ & 26.47$_{\pm 0.11}$ & 21.77 \\
            ULD \cite{boizard2024uld} \qquad & 22.80$_{\pm 0.28}$ & 15.93$_{\pm 0.74}$ & 16.43$_{\pm 0.60}$ & 26.94$_{\pm 0.28}$ & 24.83$_{\pm 0.13}$ & 20.64 \\
            % DSKD + Align (ours) & & & & & \\
            \hline
            % DSKD-CMA-KL (ours) & 25.83$_{\pm 0.43}$ & 18.62$_{\pm 0.11}$ & 18.05$_{\pm 0.72}$ & 30.78$_{\pm 0.43}$ & 30.50$_{\pm 0.20}$ & 24.76 \\
            DSKD-CMA-AKL (ours) & 26.45$_{\pm 0.56}$ & 19.57$_{\pm 0.69}$ & 17.95$_{\pm 0.55}$ & 35.99$_{\pm 0.19}$ & 35.00$_{\pm 0.16}$ & 26.99 \\
            \toprule
        \end{tabular}
    }
    \caption{Rouge-L scores (\%) on several benchmarks with TinyLLaMA-1.1B as the student. We list the mean values and the standard deviations among 5 random seeds. \modi{``\emph{w/} DSKD'' denotes our DSKD using the corresponding distance function as $\mathcal{D}(\cdot||\cdot)$ in Eqn. \eqref{eq:stuside_kd_loss}. And ``DSKD-CMA-AKL'' denotes our DSKD framework equipped with cross-model attention with AKL as $\mathcal{D}(\cdot||\cdot)$ in Eqn. \eqref{eq:stuside_kd_loss}.}}
    \label{tab:main_results_tinyllama}
    \vspace{-5pt}
\end{table*}

\subsection{Baselines}
We compare our framework with existing methods under two settings:
\paragraph{KD with the same vocabulary.} In this setting, we compare DSKD with the current white-box KD framework on the following distance functions:
\begin{itemize}
    \item \textbf{KL.} The standard KL divergence used in KD proposed by \citet{hinton15kd}.
    \item \textbf{RKL.} The reverse KL divergence that swaps the two distributions in KL divergence.
    \item \textbf{JS.} Jenson-Shannon (JS) divergence, a symmetric variant of KL divergence.
    \item \textbf{SKL.} The skewed KL proposed by \citet{ko24distillm}, which skews the student distribution $q_{\theta}$ in KL as $\lambda p + (1-\lambda)q_{\theta}$.
    \item \textbf{SRKL.} The skewed RKL proposed by \citet{ko24distillm}, which skews the teacher distribution $p$ in RKL as $\lambda q_{\theta} + (1 - \lambda) p$.
    \item \textbf{AKL.} The adaptive fusion of KL and RKL proposed by \citet{wu2024rethinking}.
\end{itemize}

\paragraph{KD with different vocabularies.} We also compare DSKD with cross-model attention to the KD methods for different vocabularies:
\begin{itemize}
    \item \textbf{MinCE.} The method proposed by \citet{wan24fusellm}, aligns the logits between different models via dynamic programming that minimizes the edit distances of token strings.
    \item \textbf{ULD.} The method proposed by \citet{boizard2024uld}, replaces the usual KL divergence with a closed-form solution of Wasserstein distance to overcome the limitation on the same tokenizers between the teacher and the student.
\end{itemize}

Besides, we also compare our framework with \modi{the black-box KD method, \emph{i.e.},} sequence-level KD (\textbf{SeqKD}) \cite{kim16seqkd}, under both settings.
Nevertheless, we did not compare our framework with on-policy KD methods such as ImitKD \cite{lin20imitkd}, GKD \cite{agarwal24gkd}, MiniLLM \cite{gu23minillm} and DistiLLM \cite{ko24distillm} since we only focus on the more general off-policy scenarios.

\subsection{Results}

\paragraph{KD with the same vocabulary.} 
The results of KD for models with the same vocabulary are presented at the top parts of Table \ref{tab:main_results_gpt2} and Table \ref{tab:main_results_tinyllama}.
Firstly, it is shown that all white-box KD methods exhibit better performance than the black-box KD method SeqKD, which demonstrates that token-level distributions can transfer more knowledge than single target tokens.
Furthermore, our DSKD framework significantly outperforms the current white-box KD framework for both GPT2 and TinyLLaMA on various distance functions.
On the one hand, it showcases the effectiveness of our DSKD framework that conducts KD in unified output spaces.
On the other hand, the improvements on all distance functions also demonstrate that our framework is highly compatible with current distance functions in KD.

\begin{table}[h]
    \centering
    \resizebox{\linewidth}{!}{
    \begin{tabular}{lccc}
        \bottomrule
        \textbf{Objective} & \textbf{Diff. Space} & \textbf{Student Space} & \textbf{DSKD} \\
        \hline
        \hline
        \multicolumn{4}{c}{\textbf{GPT2-1.5B $\rightarrow$ GPT2-120M}} \\
        \hline
        KL & 17.63 & 18.00 & 18.69 \\
        RKL & 17.82 & 18.03 & 18.44 \\
        JS & 16.98 & 17.17 & 18.37 \\
        SKL & 17.71 & 17.99 & 18.14 \\
        SRKL & 17.19 & 17.47 & 18.29 \\
        AKL & 17.63 & 17.77 & 18.39 \\
        \hline
        \multicolumn{4}{c}{\textbf{LLaMA2-7B $\rightarrow$ TinyLLaMA-1.1B}} \\
        \hline
        KL & 23.53 & 24.99 & 25.05 \\
        RKL & 23.47 & 25.50 & 25.63 \\
        JS & 22.62 & 22.64 & 23.44 \\
        SKL & 23.01 & 23.55 & 24.17 \\
        SRKL & 23.25 & 23.64 & 24.30 \\
        AKL & 23.30 & 26.23 & 26.65 \\
        \toprule
    \end{tabular}
    }
    \caption{The averaged Rouge-L (\%) among all test sets. The detailed scores on each test set are in Appendix \ref{sec:full_ablation}.}
    \label{tab:ablation}
    \vspace{-5pt}
\end{table}

\paragraph{KD with different vocabularies.} 
At the bottom parts of Table \ref{tab:main_results_gpt2} and Table \ref{tab:main_results_tinyllama}, we also show the results of KD methods for models with different vocabularies\footnote{In this setting, we only list the results of our method with the best performing distance functions due to space limitation. The full results are listed in Table \ref{tab:full_results_gpt2} and Table \ref{tab:full_results_tinyllama}.}.
As mentioned in \S\ref{sec:depend_same_vocab}, the key challenge in this setting is to deal with the mismatch distributions due to different vocabulary sizes and tokenization.
Facing this challenge, existing KD methods only pre-define coarse alignment and thus yield limited performance, lagging behind KD methods for models with the same vocabulary.
In contrast, our CMA mechanism learns the alignment automatically, with which our DSKD performs better than existing methods.
Particularly, as the teacher models under this setting are stronger, DSKD-CMA can sometimes achieve better performance than DSKD with the same vocabulary (\emph{e.g.}, DSKD-CMA-AKL in Table \ref{tab:main_results_tinyllama}).
It suggests the potential of our method to train better students with stronger teachers, even if they have different vocabularies.

\section{Analysis}
% \vspace{-10pt}
\subsection{KD in Different Spaces \emph{vs.} Unified Space}
In this section, we further evaluate whether unifying the space for KD leads to better performance.
Specifically, we only keep the KD process in the student space in our DSKD, \emph{i.e.}, only calculate the losses in Eqn. \eqref{eq:stuside_ce_loss} and Eqn. \eqref{eq:stuside_kd_loss}, since it optimizes the same student distribution $q_{\theta}$ as the current KD framework does in Eqn. \eqref{eq:kd_loss}.
The only difference is that the teacher distribution $p^{t \rightarrow s}$ in Eqn. \eqref{eq:stuside_kd_loss} shares the same output space with the student distribution. 
The results are shown in Table \ref{tab:ablation}.
For all distance functions, KD in the student space (\textbf{Student Space}) consistently surpasses KD in different spaces (\textbf{Diff. Space}).
These results sufficiently reflect the superiority of unifying the output spaces of the distributions for KD.
Furthermore, when combined with KD in the teacher space, KD in dual spaces, \emph{i.e.}, DSKD, achieves further improvement, indicating that KD in the student space and the teacher space can complement each other.

\subsection{Evaluation via GPT-4}
\label{sec:llm_eval}
\begin{figure}
    \centering
    \resizebox{\linewidth}{!}{
        \includegraphics{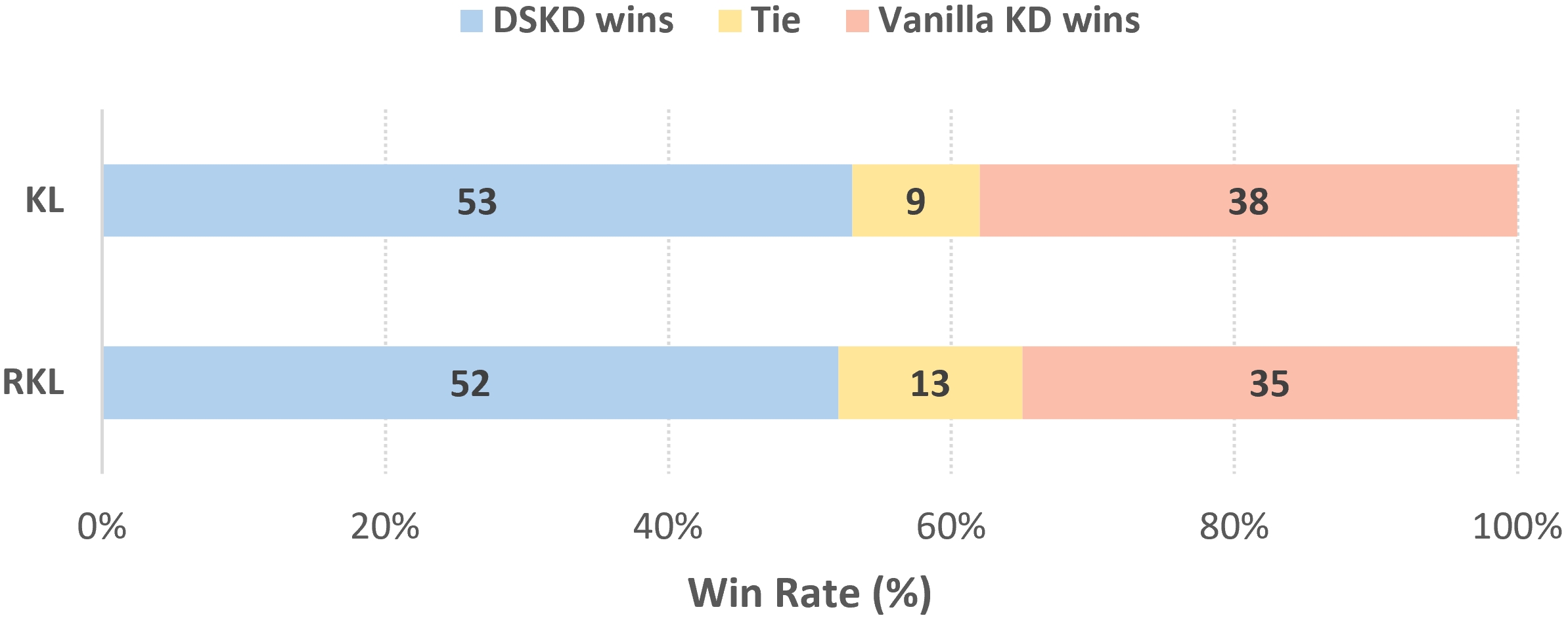}
    }
    \caption{Win rates (\%) on the response quality between TinyLLaMA trained by DSKD and the current white-box KD framework.}
    \label{fig:llm_eval}
\end{figure}

We also use GPT-4 to evaluate and compare our DSKD and the current white-box KD framework.
Specifically, we randomly pick 100 instructions in the test set of Dolly and generate responses with TinyLLaMA trained by DSKD and the current framework.
Then we use GPT-4 to judge which responses are better and plot the win rates in Figure \ref{fig:llm_eval}.
It is shown that our DSKD can beat the current KD framework in most cases for both KL divergence and reverse KL divergence.
More details and the complete results for other distance functions can be referred to in Appendix \ref{sec:llm_eval_full}.

\subsection{Representation Similarity between the Teacher and the Student}
\begin{figure}
	\centering
	\subfigure[Cosine as Structure]{
		\begin{minipage}[t]{0.47\linewidth}
			\centering
			\includegraphics[width=\linewidth]{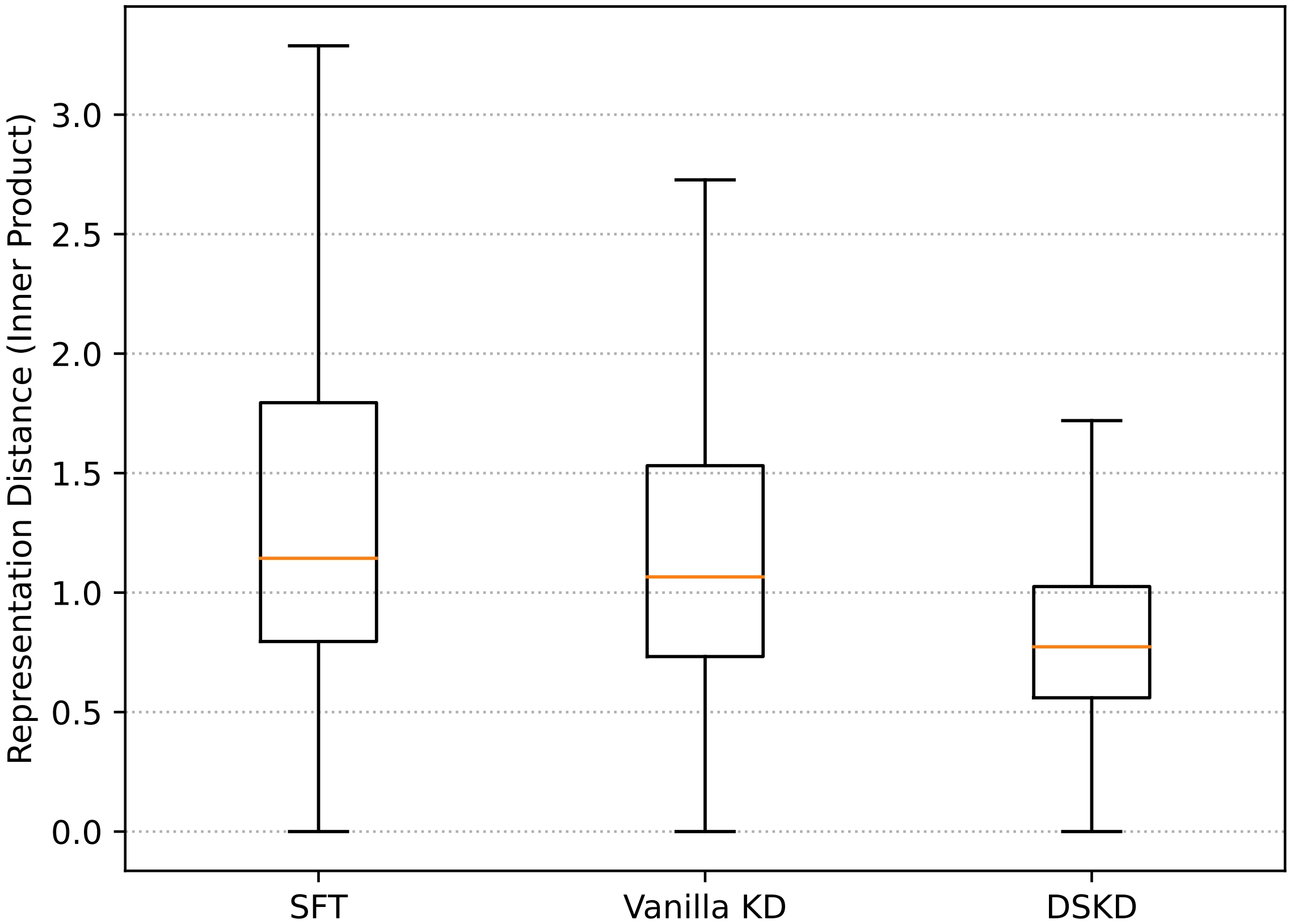}
		\end{minipage}
	}%
	\subfigure[Inner Product as Structure]{
		\begin{minipage}[t]{0.47\linewidth}
			\centering
			\includegraphics[width=\linewidth]{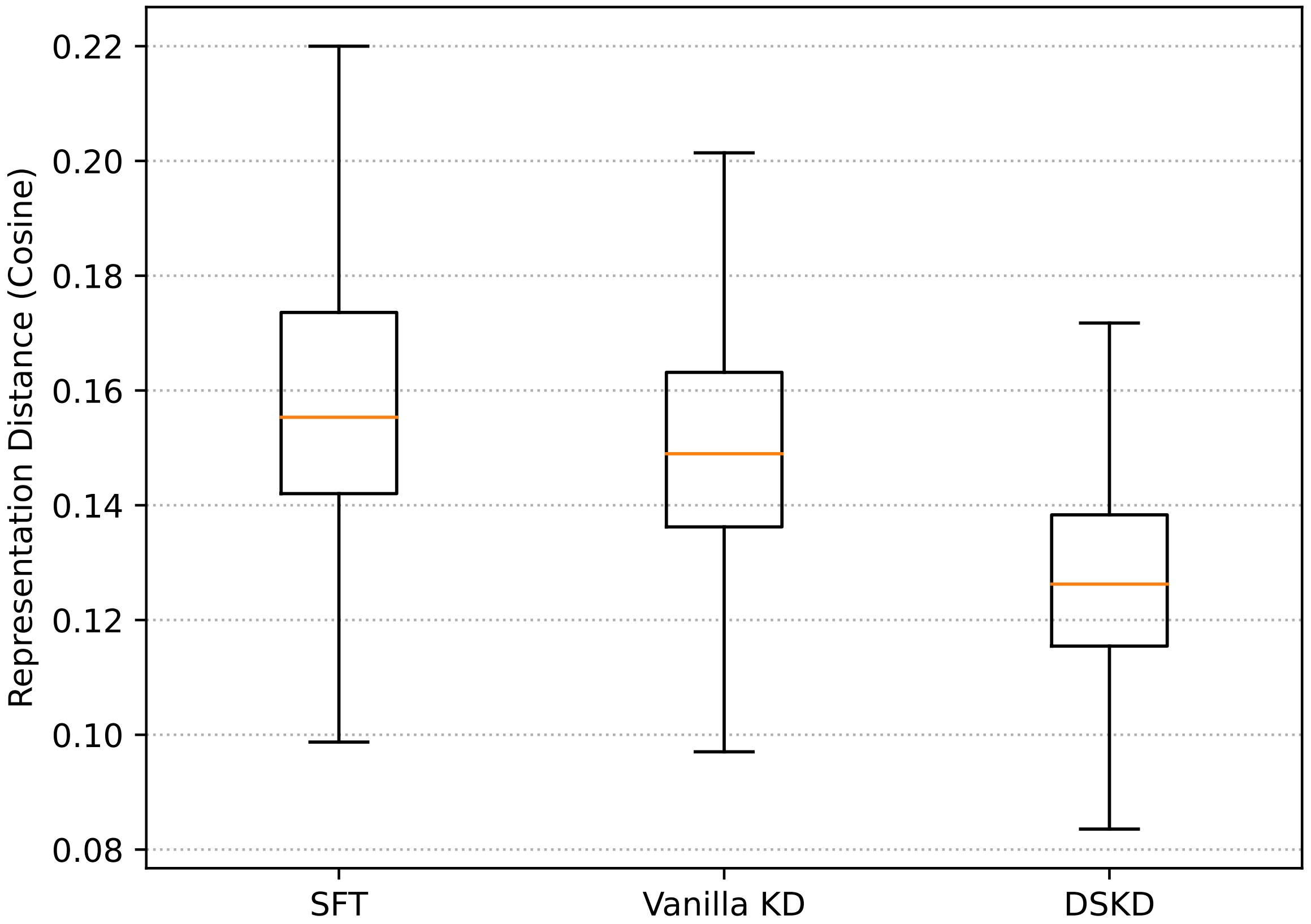}
		\end{minipage}
	}%
	\centering
	\caption{Distance between the representation structures of the teacher and the student.}
	\label{fig:repr_sim}
\end{figure}

In the simulation experiment, we find that the current KD framework will lead to limited representation similarities between the student and the teacher (as shown in Figure \ref{fig:kl_simulation}(b)). 
Thus, we evaluate whether this phenomenon also holds in the real KD scenario.
Since the dimensions are usually different for the teacher and student models, we measure the similarity of representation structures of the two models instead of their hidden states.
Specifically, we use cosine similarity and normalized inner product between output hidden states to represent the representation structure of a model (see Eqn. \eqref{eq:cosine} and \eqref{eq:inner_prod} in Appendix \ref{sec:repr_distance} for the definitions). 
Then we calculate the L1 distance between the representation structures to reflect their similarity, where lower distance denotes higher similarity between representation structures (see Eqn. \eqref{eq:d_cosine} and \eqref{eq:d_inner_prod} in Appendix \ref{sec:repr_distance} for the detailed calculations).
The average distances between the structure of the teacher and the student on 1000 training samples are plotted in Figure \ref{fig:repr_sim}.
It shows that on both types of representation structures, the current KD framework (\textbf{Vanilla KD}) only reduces minor distances between the teacher and the student compared to fine-tuning without KD (\textbf{SFT}).
However, our DSKD achieves significantly lower distances between the teacher and the student, which indicates that DSKD can enhance the similarity between the student and the teacher.

% \subsection{Alignment Map in Cross-Model Attention}

\section{Related Work}
% \subsection{White-Box KD for Language Models}
\paragraph{White-Box KD for Language Models.}
The white-box KD framework for language models stems from the standard KD method proposed by \citet{hinton15kd}.
As pre-trained language models (PLMs) become prevalent for various NLP tasks, numerous KD methods within this framework were proposed to compress the excessive model sizes of PLMs \cite{sun19patientkd,sanh19distilbert,sun20mobilebert,jiao20tinybert}.
Besides minimizing the distance between distributions, there are also feature-based KD methods that distill the knowledge in intermediate hidden states and attention maps of the teacher model \cite{jiao20tinybert,wang20minilm,wang21minilmv2}.
% Also, white-box KD is widely used in text generation tasks, such as neural machine translation \cite{tan19mnmtkd}, text abstraction\cite{chen19bertkd4textgen}, and so on.
Additionally, white-box KD is also widely used in text generation tasks, such as neural machine translation \cite{tan19mnmtkd,wang-etal-2021-selective,zhang-etal-2023-towards-understanding} and text summarization \cite{chen-etal-2020-distilling,liu-etal-2021-noisy}.
Since LLMs are predominate for various tasks, several KD techniques have also been proposed for LLMs \cite{gu23minillm,ko24distillm,wu2024rethinking,xu24llmkdsurvey}.
Unlike the previous work that follows the current white-box KD framework, we challenge this framework by revealing its inherent limitations and proposing a simple yet more effective and general KD framework as the solution.
\paragraph{KD with the Shared Prediction Head.}
In the previous literature on KD, SimKD \citep{chen22simkd} also proposed to share the teacher's prediction head for KD, which was similar to the process of KD in the teacher space in our DSKD.
However, the aim of SimKD is to equip the prediction head of the teacher model to the student model, and thus the student model will be larger after KD and suffer from higher inference costs.
In contrast, our DSKD only leverages this process to transfer the representation information from the teacher and has no influence on the original model size of the student.

\section{Conclusion}
In this work, we first reveal two limitations in the current white-box KD framework for LLMs, \emph{i.e.}, leading to low similarity between the student and the teacher and the requirements of the same vocabulary between two LLMs.
To address them, we propose a novel white-box KD framework, named dual-space knowledge distillation (DSKD), which unifies the output spaces of the student and the teacher for KD.
% Our framework is also compatible with various objectives for the current white-box KD framework.
On this basis, we further develop a cross-model attention mechanism to solve the vocabulary mismatch between different LLMs, so that our DSKD framework supports KD between any two LLMs, regardless of their vocabularies.
Experimental results on several instruction-following benchmarks showcase that our framework significantly outperforms the current white-box KD framework on various distance functions.
Meanwhile, for LLMs with different vocabularies, DSKD also surpasses all existing KD methods.
% We believe our framework has the potential to be a better alternative than the current white-box KD framework for the era of LLM.

\section*{Limitations}
Although our DSKD supports KD between LLMs with different vocabularies via the cross-model attention mechanism, the final performance of DSKD-CMA in most cases still lags slightly behind the performance of DSKD when LLMs have the same vocabularies (see Table \ref{tab:full_results_gpt2} and Table \ref{tab:full_results_tinyllama}).
We attribute this gap to the alignment error between the tokens in two differently tokenized sequences.
Nevertheless, we still believe that our cross-model attention is a simple yet relatively effective method to solve the KD for LLMs with different vocabularies and may inspire more effective methods in future work.
% More advanced methods may be explored in future work.

\section*{Acknowledgements}
The research work described in this paper has been supported by the National Nature Science Foundation of China (No. 62476023, 61976016, 62376019, 61976015), and the authors would like to thank the anonymous reviewers for their valuable comments and suggestions to improve this paper.

\bibliography{custom}

\appendix
\onecolumn
\newpage
\section{Appendix}
\label{sec:appendix}

\subsection{Simulation Results for Other Distance Functions}
\label{sec:other_simulation}
\begin{figure}[h]
	\centering
	\subfigure[Before KD]{
		\begin{minipage}[t]{0.24\linewidth}
			\centering
			\includegraphics[width=\linewidth]{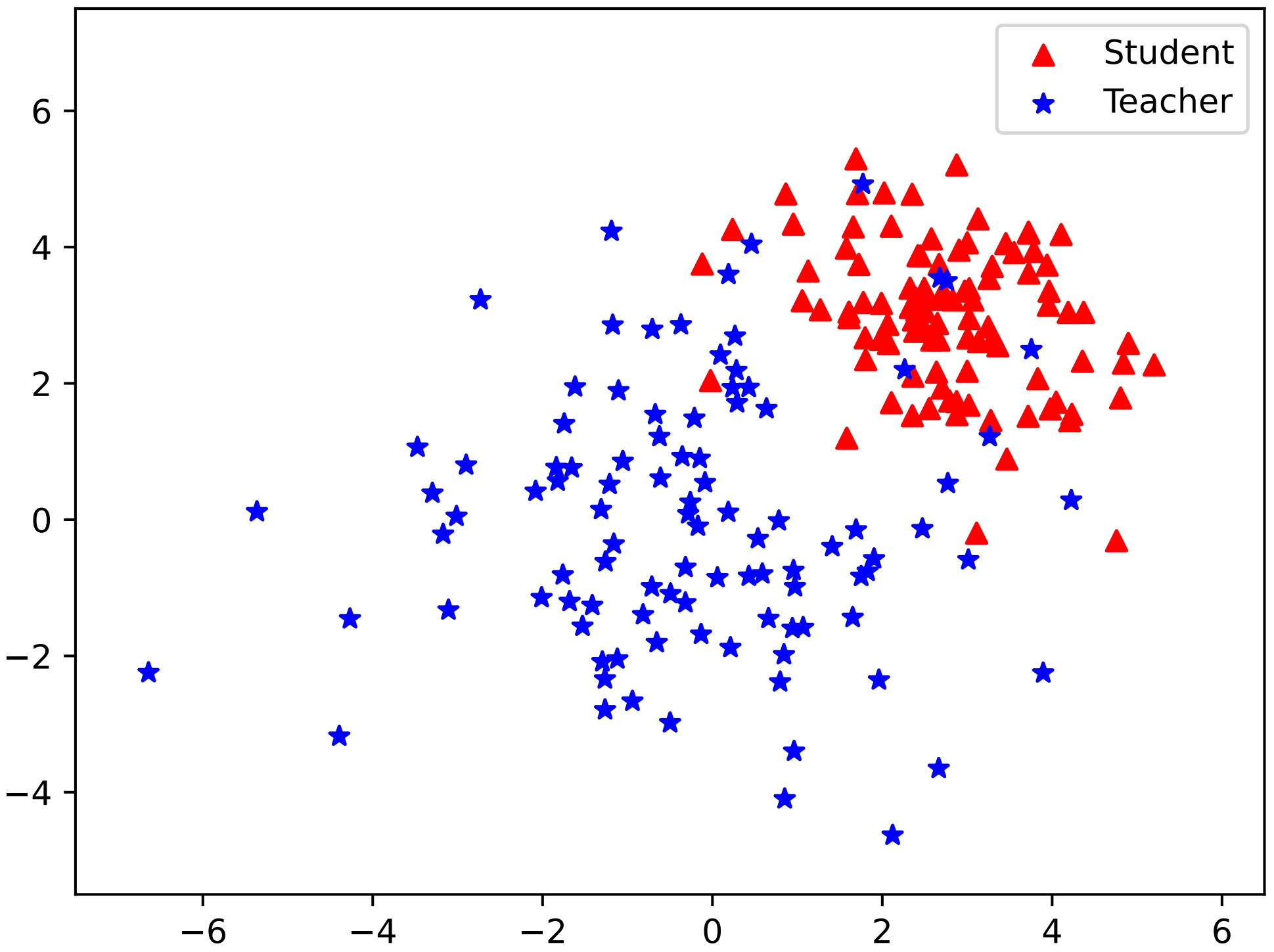}
		\end{minipage}
	}%
	\subfigure[After KD (different heads)]{
		\begin{minipage}[t]{0.24\linewidth}
			\centering
			\includegraphics[width=\linewidth]{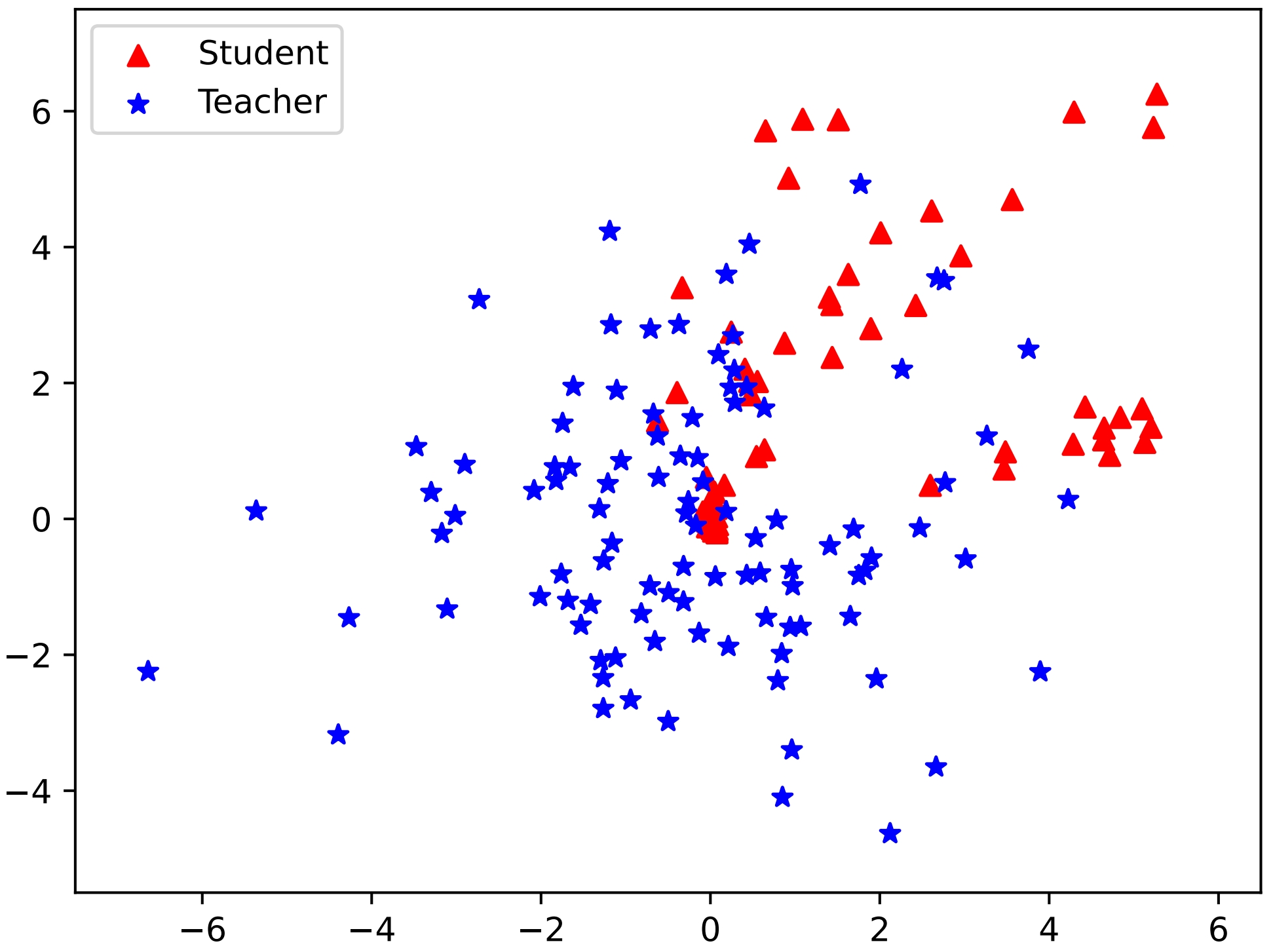}
		\end{minipage}
	}%
	%此处的空行很重要，想让图片在什么地方换行就在代码对应位置空行
	\subfigure[After KD (shared head)]{
		\begin{minipage}[t]{0.24\linewidth}
			\centering
			\includegraphics[width=\linewidth]{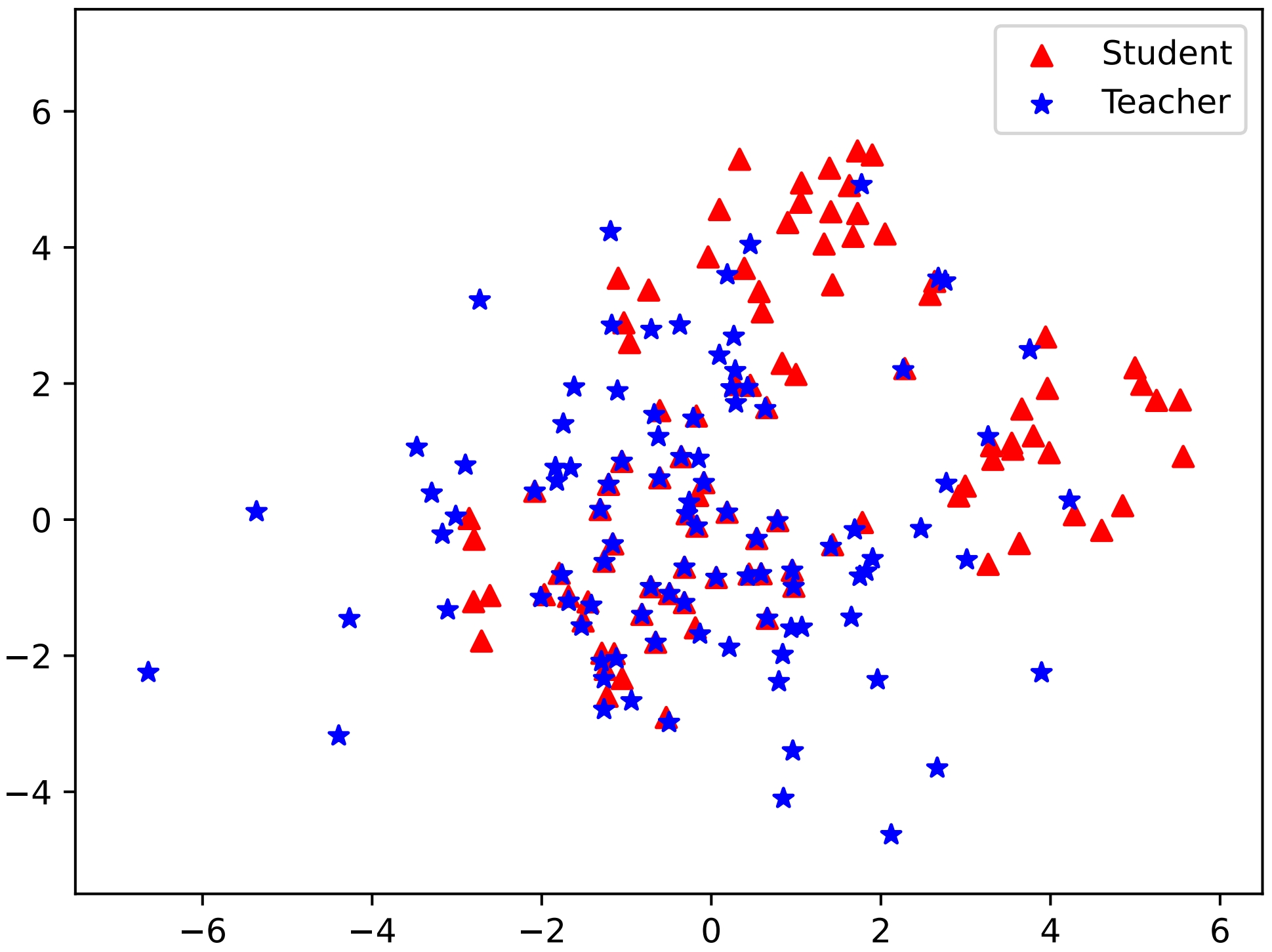}
		\end{minipage}
	}%
	\subfigure[Loss curves of KD]{
		\begin{minipage}[t]{0.24\linewidth}
			\centering
			\includegraphics[width=\linewidth]{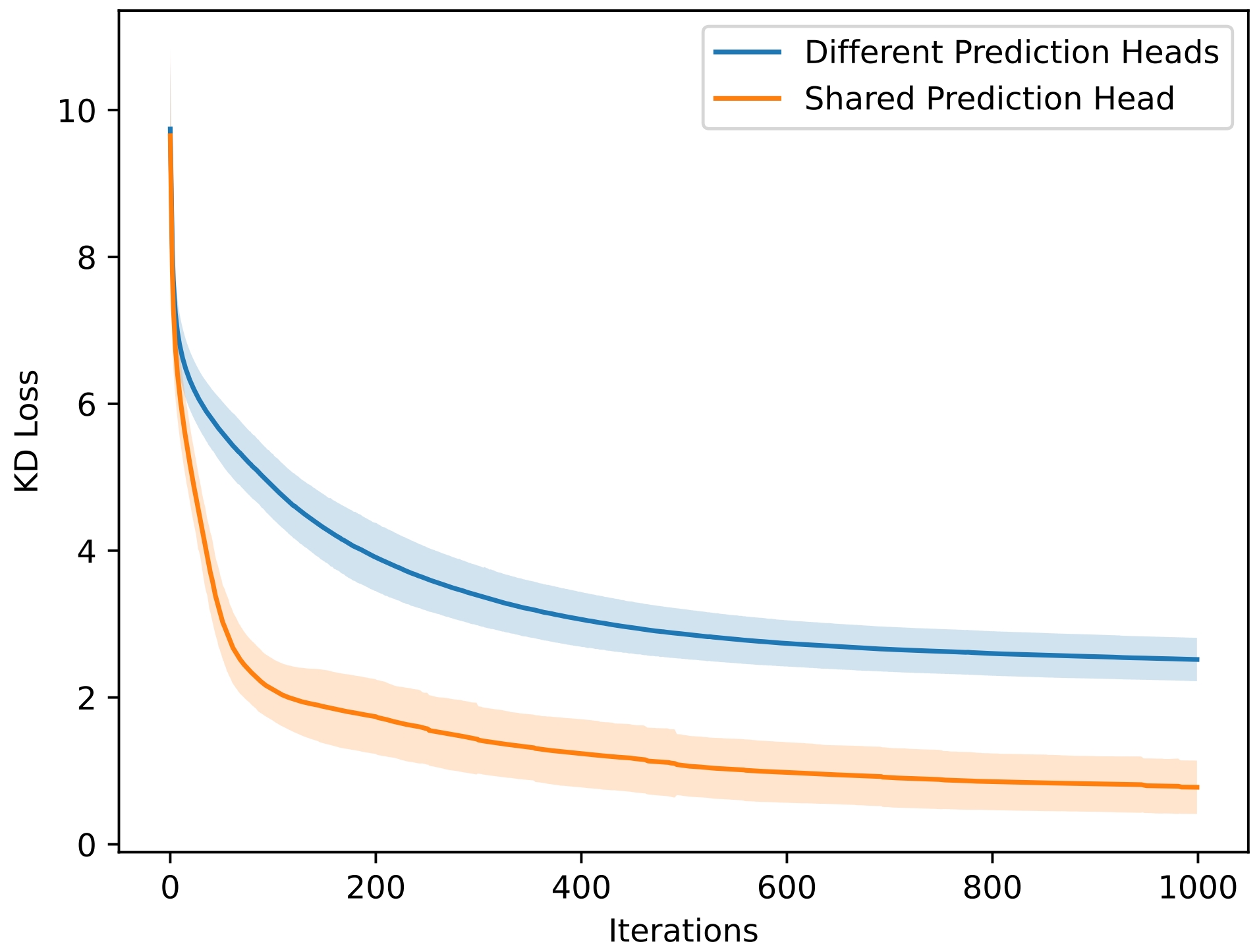}
		\end{minipage}
	}%
	\centering
	\caption{Simulation results with reverse KL divergence as the KD objective. (a), (b) and (c) plot the {\color{red} student's hidden states} and the {\color{blue} teacher's hidden states} before and after the two KD processes. (d) shows the convergence curves of the KD objective in the two KD processes.}
	\label{fig:rkl_simulation}
\end{figure}

\begin{figure}[h]
	\centering
	\subfigure[Before KD]{
		\begin{minipage}[t]{0.24\linewidth}
			\centering
			\includegraphics[width=\linewidth]{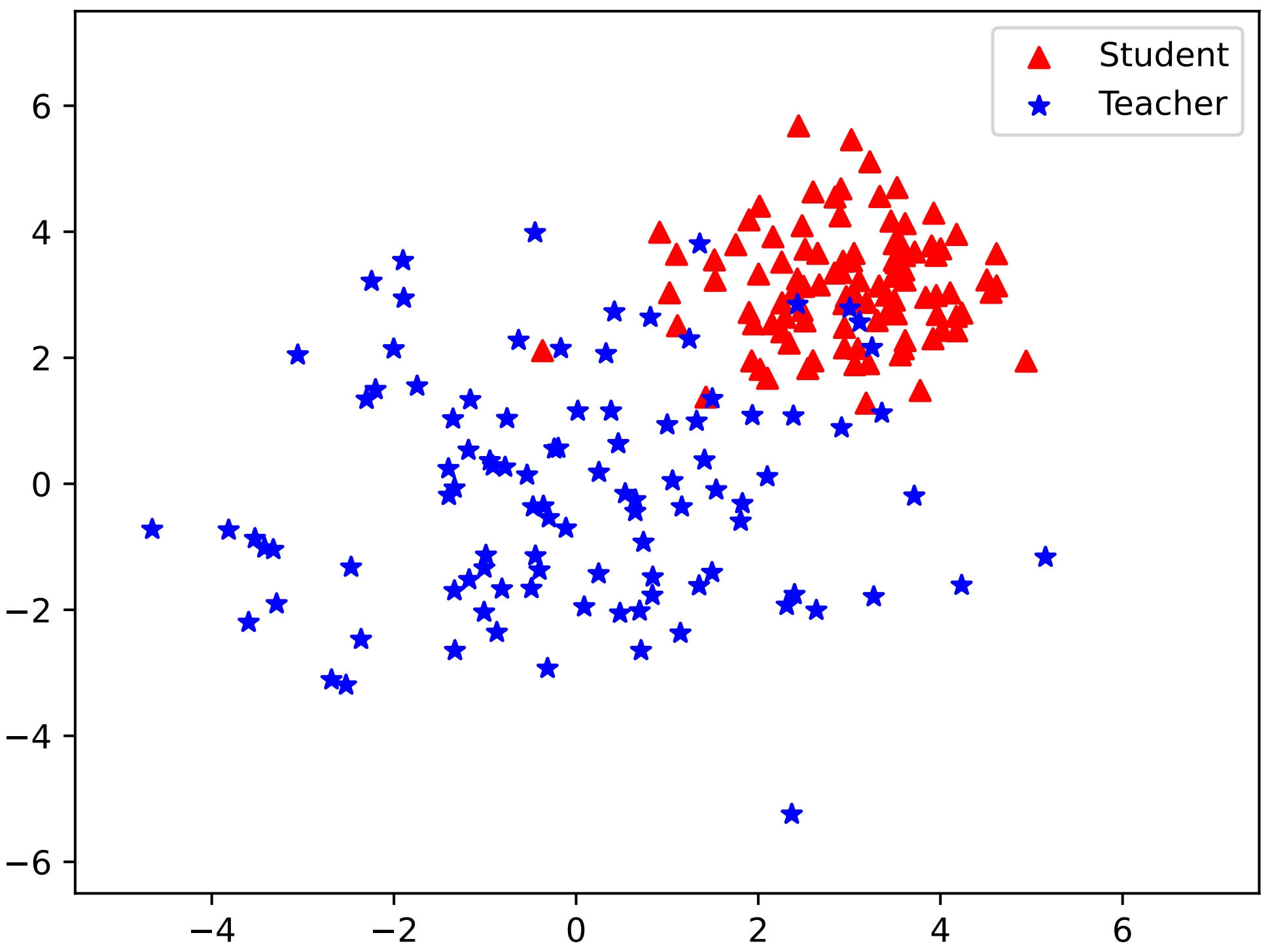}
		\end{minipage}
	}%
	\subfigure[After KD (different heads)]{
		\begin{minipage}[t]{0.24\linewidth}
			\centering
			\includegraphics[width=\linewidth]{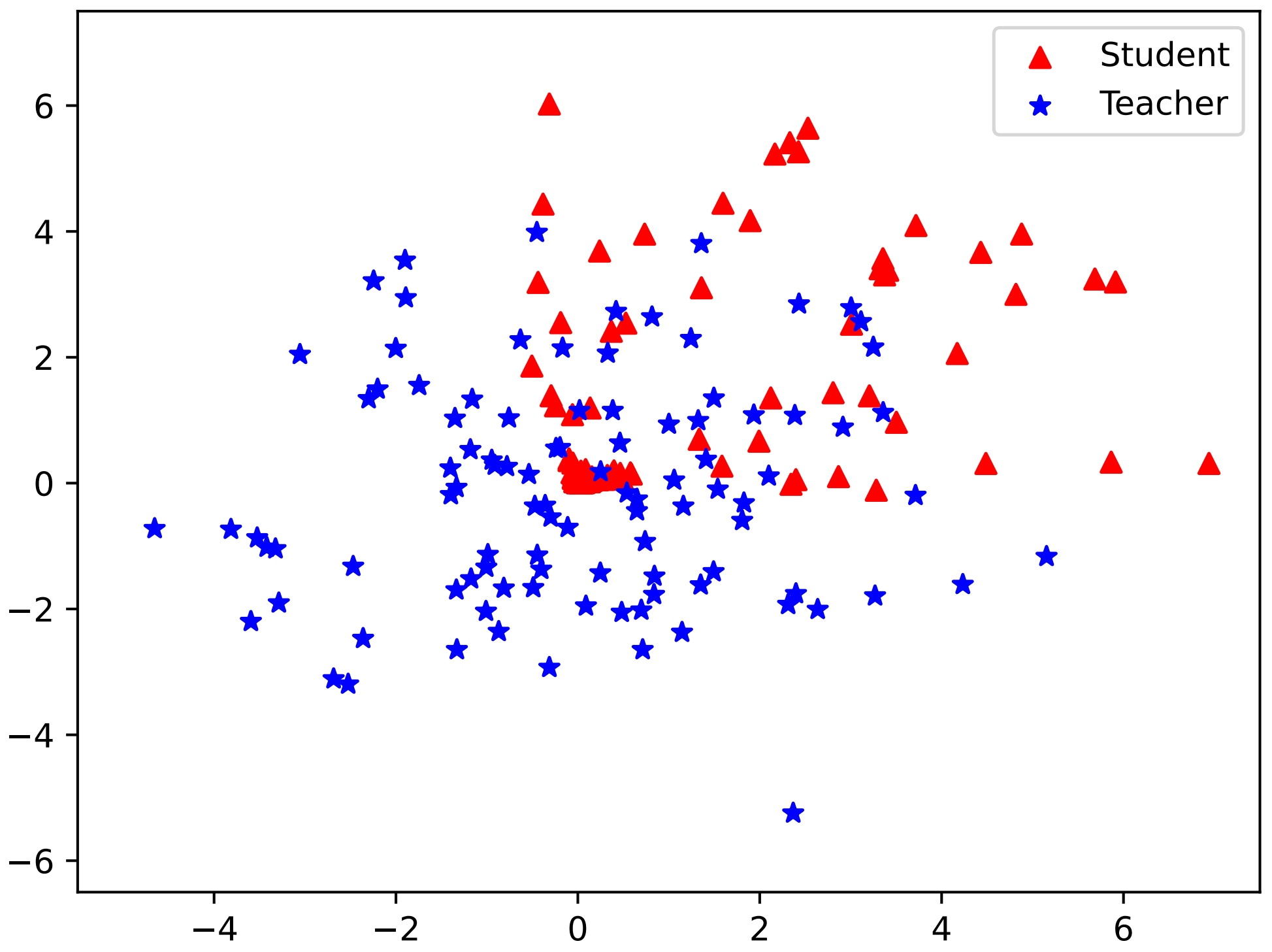}
		\end{minipage}
	}%
	%此处的空行很重要，想让图片在什么地方换行就在代码对应位置空行
	\subfigure[After KD (shared head)]{
		\begin{minipage}[t]{0.24\linewidth}
			\centering
			\includegraphics[width=\linewidth]{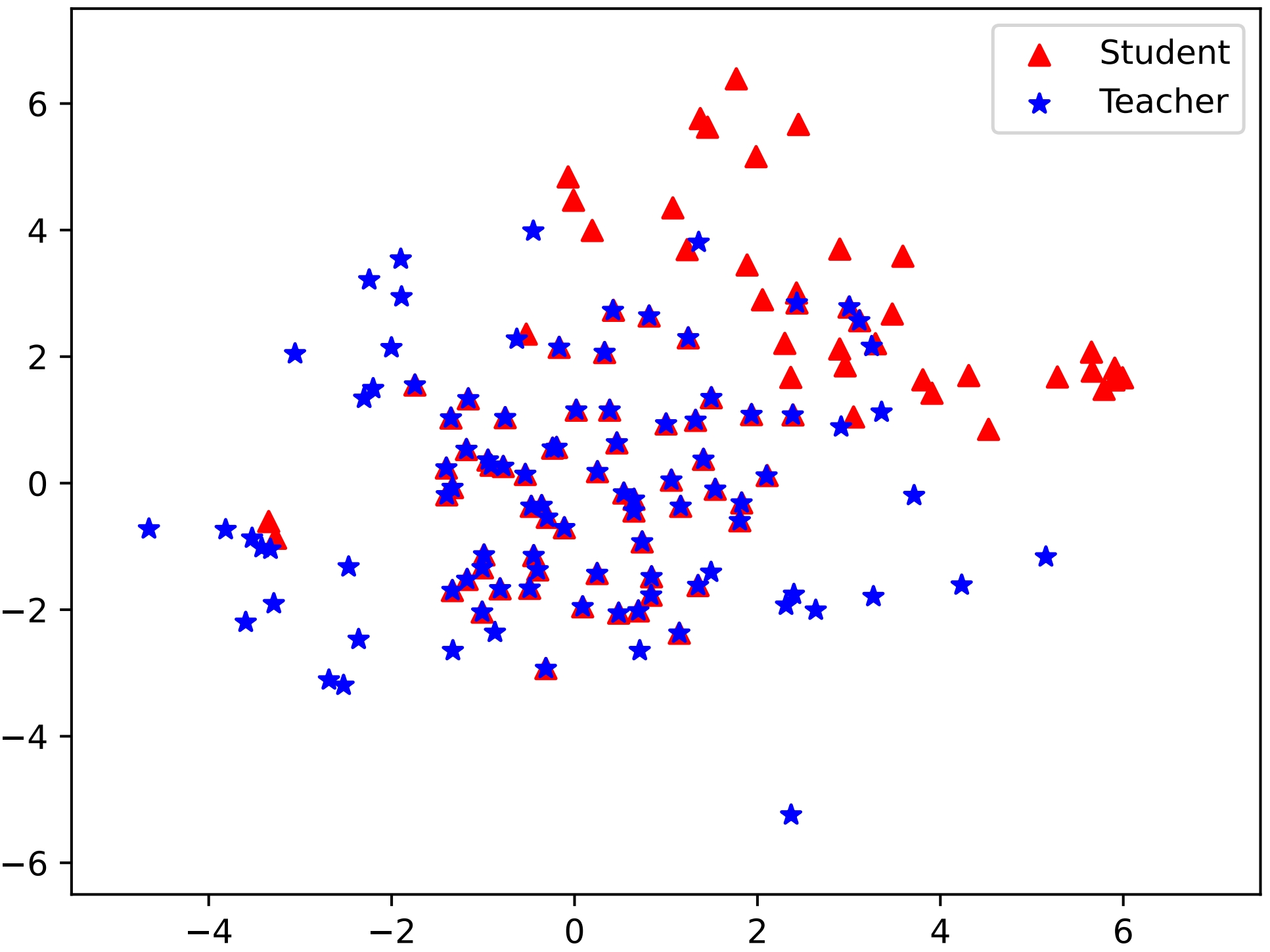}
		\end{minipage}
	}%
	\subfigure[Loss curves of KD]{
		\begin{minipage}[t]{0.24\linewidth}
			\centering
			\includegraphics[width=\linewidth]{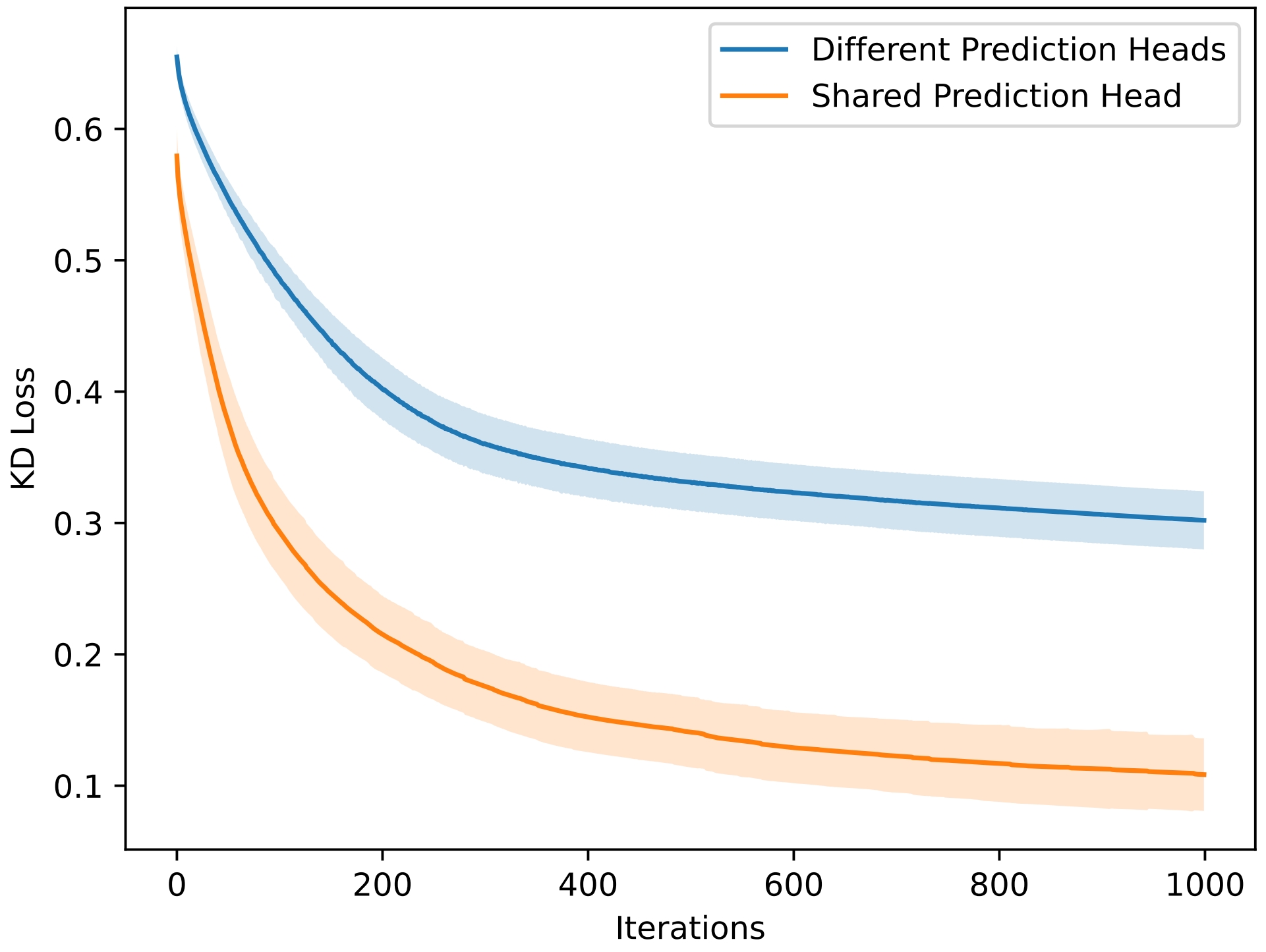}
		\end{minipage}
	}%
	\centering
	\caption{Simulation results with JS divergence as the KD objective. (a), (b) and (c) plot the {\color{red} student's hidden states} and the {\color{blue} teacher's hidden states} before and after the two KD processes. (d) shows the convergence curves of the KD objective in the two KD processes.}
	\label{fig:js_simulation}
\end{figure}

\begin{figure}[h]
	\centering
	\subfigure[Before KD]{
		\begin{minipage}[t]{0.24\linewidth}
			\centering
			\includegraphics[width=\linewidth]{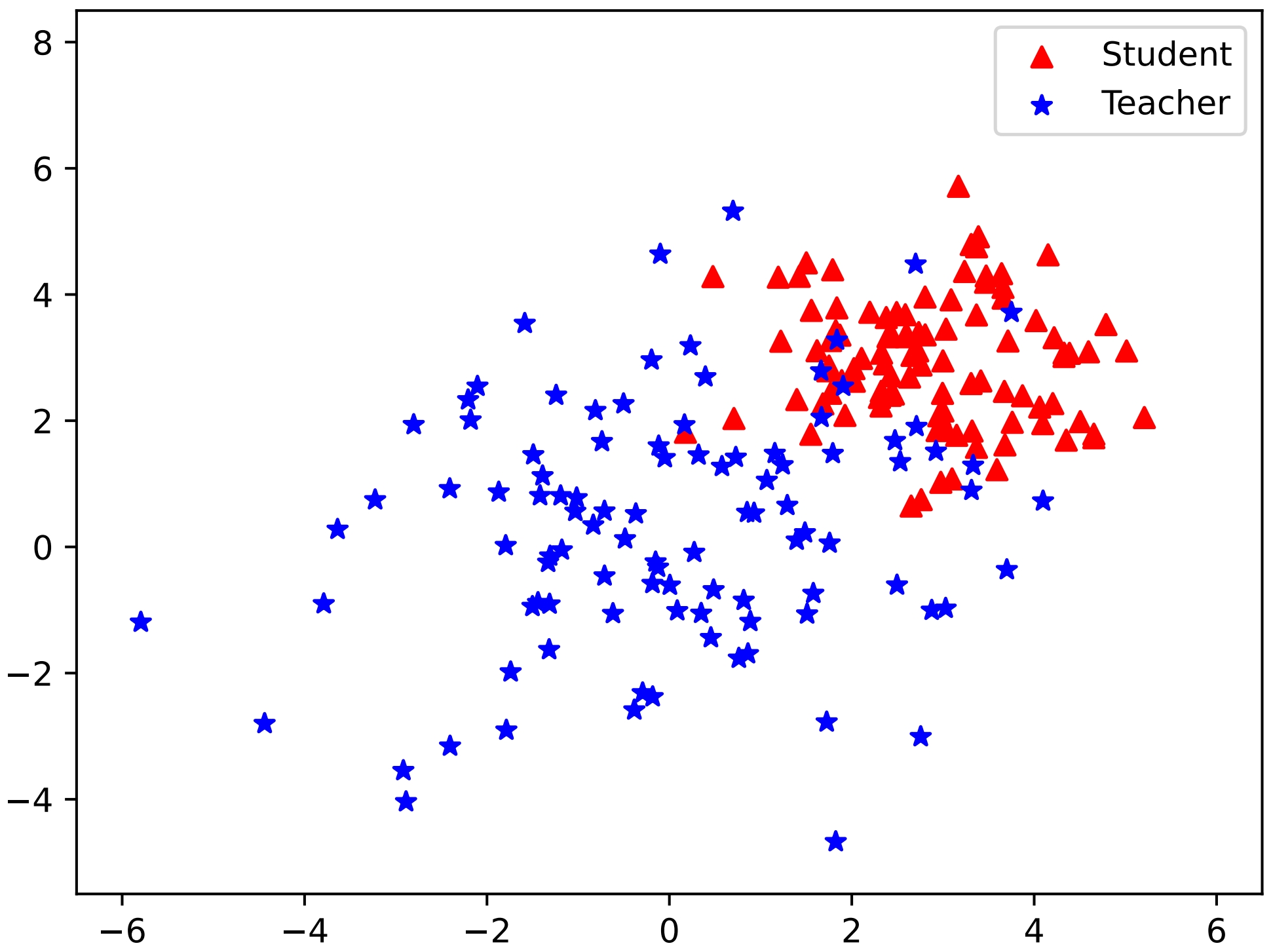}
		\end{minipage}
	}%
	\subfigure[After KD (different heads)]{
		\begin{minipage}[t]{0.24\linewidth}
			\centering
			\includegraphics[width=\linewidth]{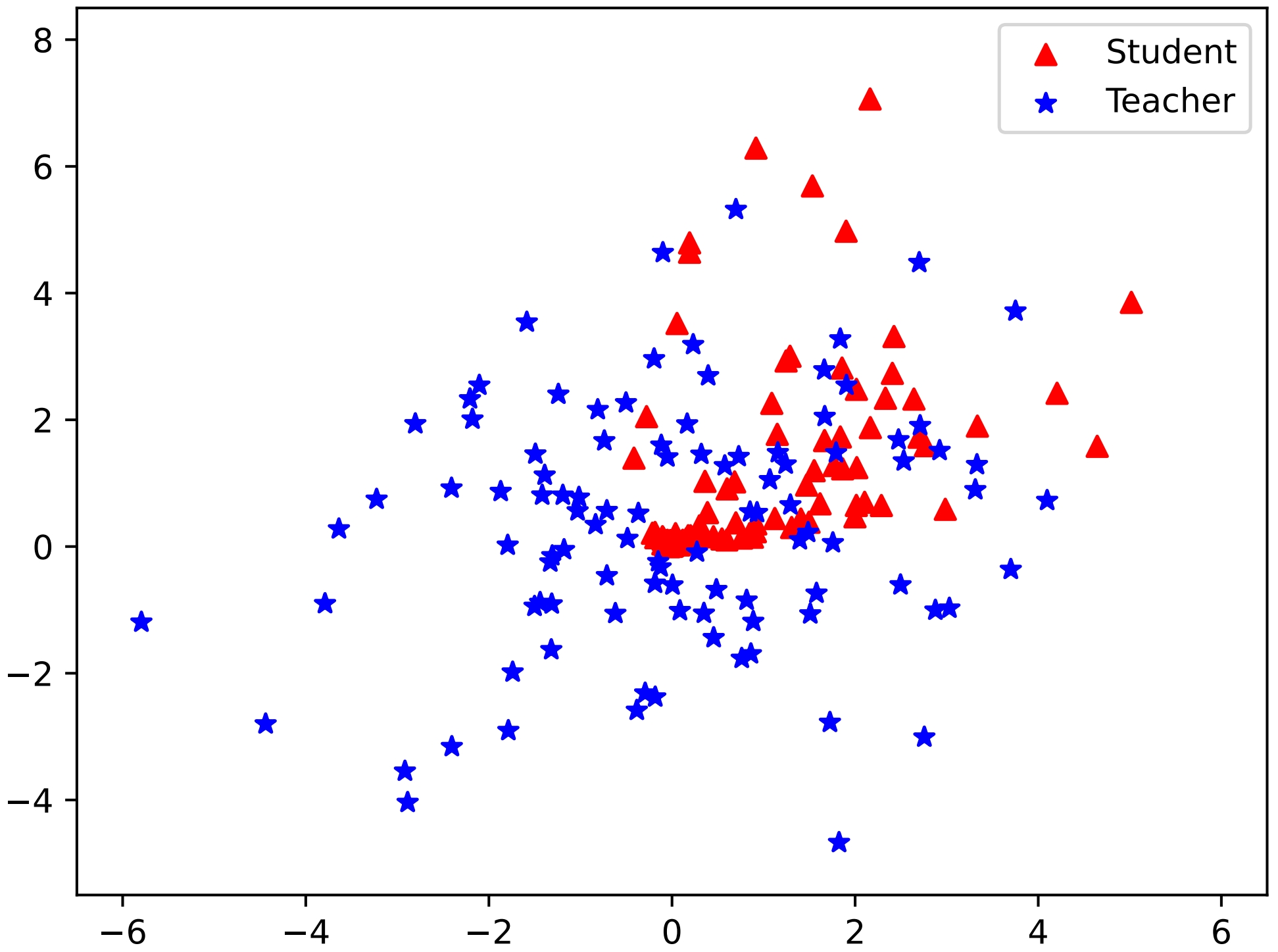}
		\end{minipage}
	}%
	%此处的空行很重要，想让图片在什么地方换行就在代码对应位置空行
	\subfigure[After KD (shared head)]{
		\begin{minipage}[t]{0.24\linewidth}
			\centering
			\includegraphics[width=\linewidth]{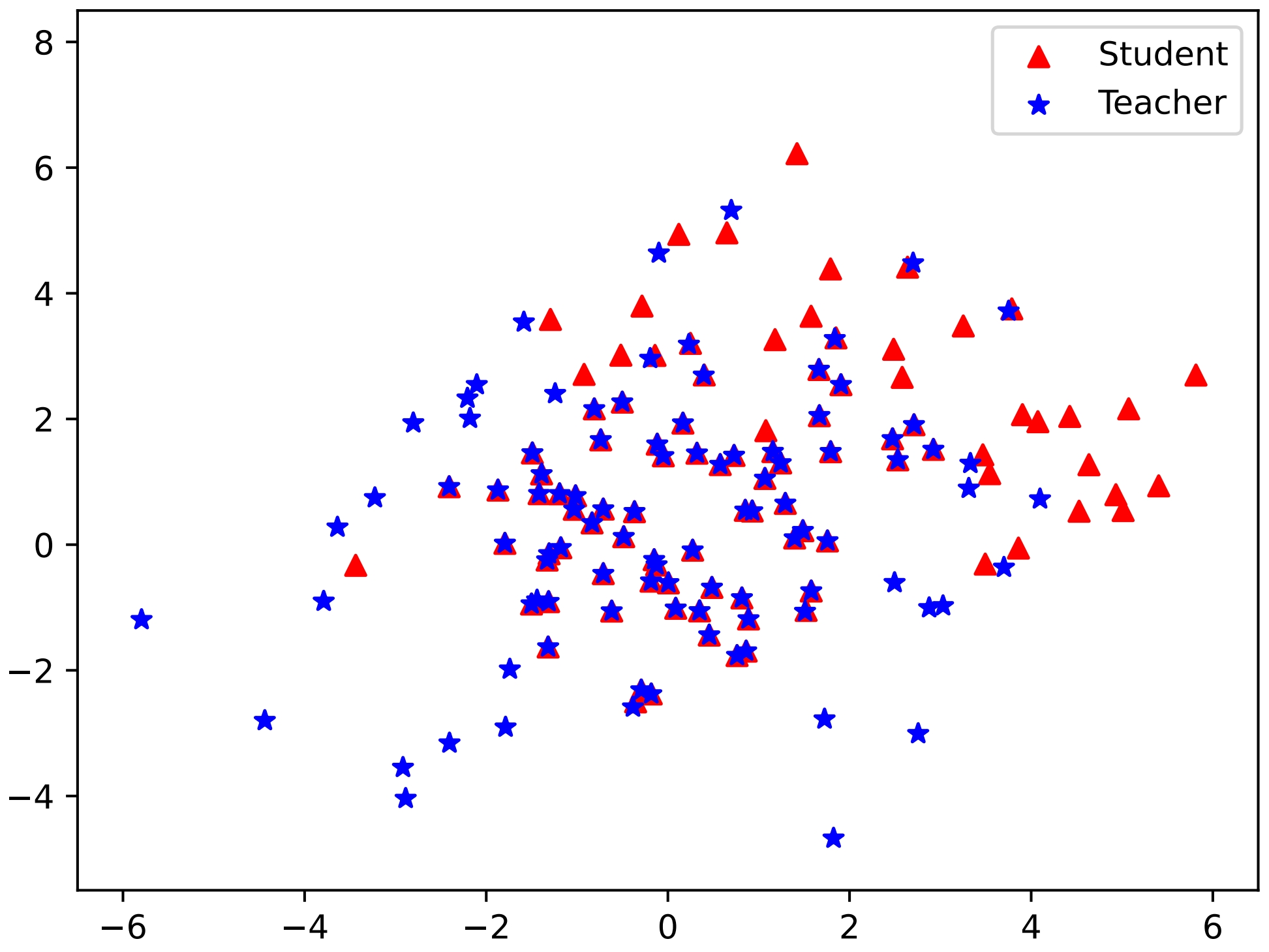}
		\end{minipage}
	}%
	\subfigure[Loss curves of KD]{
		\begin{minipage}[t]{0.24\linewidth}
			\centering
			\includegraphics[width=\linewidth]{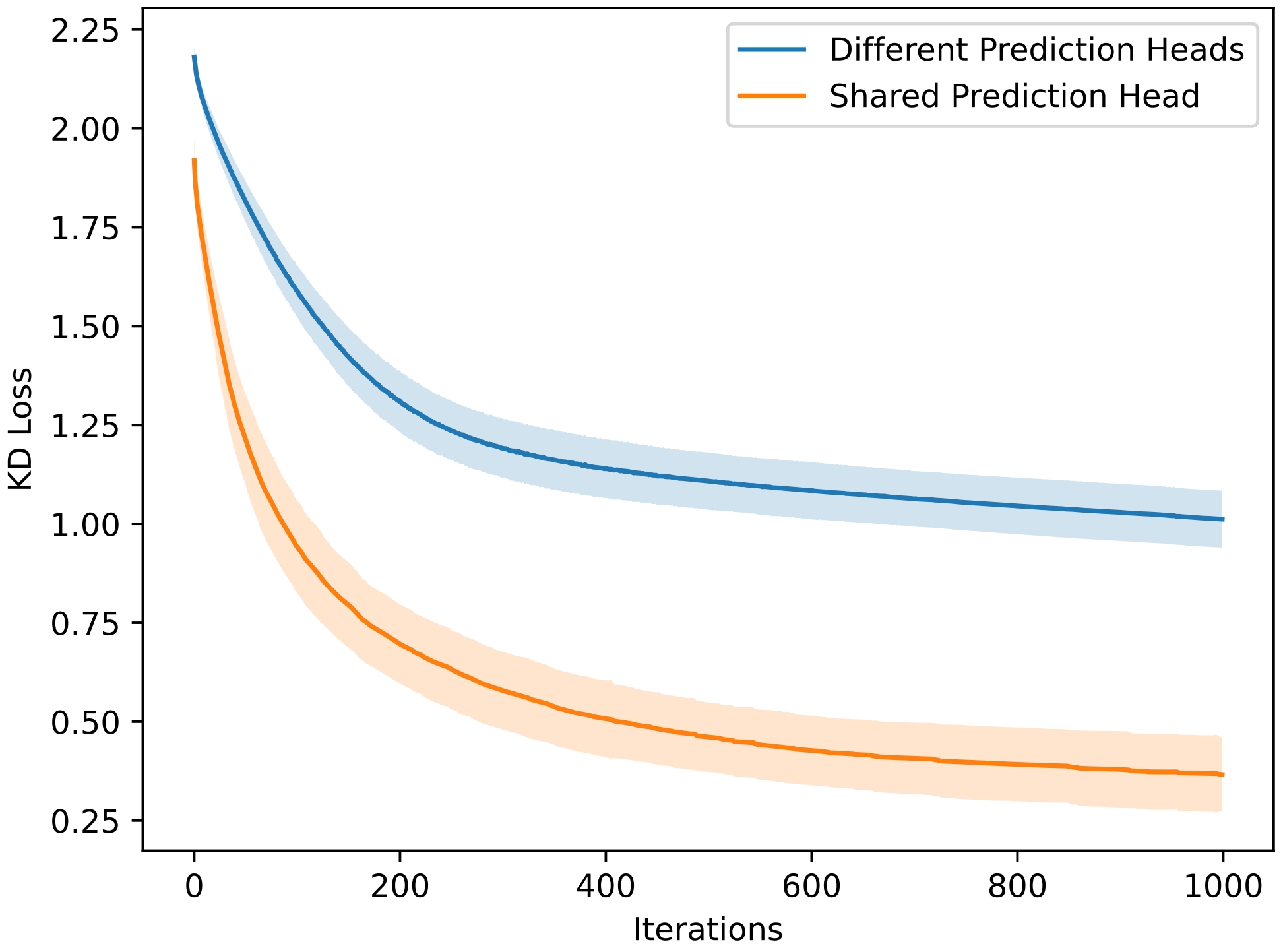}
		\end{minipage}
	}%
	\centering
	\caption{Simulation results with skewed KL divergence as the KD objective. (a), (b) and (c) plot the {\color{red} student's hidden states} and the {\color{blue} teacher's hidden states} before and after the two KD processes. (d) shows the convergence curves of the KD objective in the two KD processes.}
	\label{fig:skl_simulation}
\end{figure}

\begin{figure}[t]
	\centering
	\subfigure[Before KD]{
		\begin{minipage}[t]{0.24\linewidth}
			\centering
			\includegraphics[width=\linewidth]{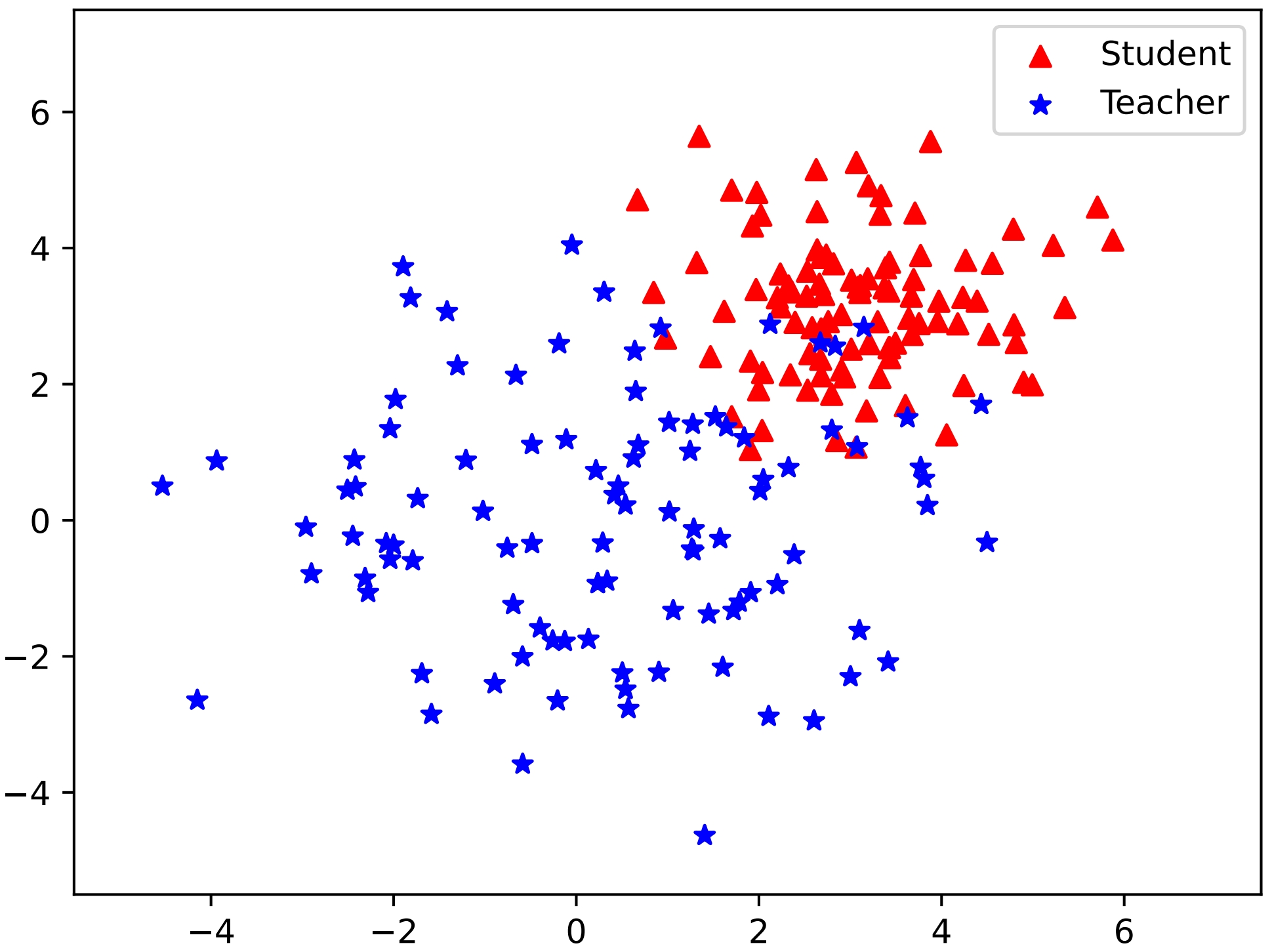}
		\end{minipage}
	}%
	\subfigure[After KD (different heads)]{
		\begin{minipage}[t]{0.24\linewidth}
			\centering
			\includegraphics[width=\linewidth]{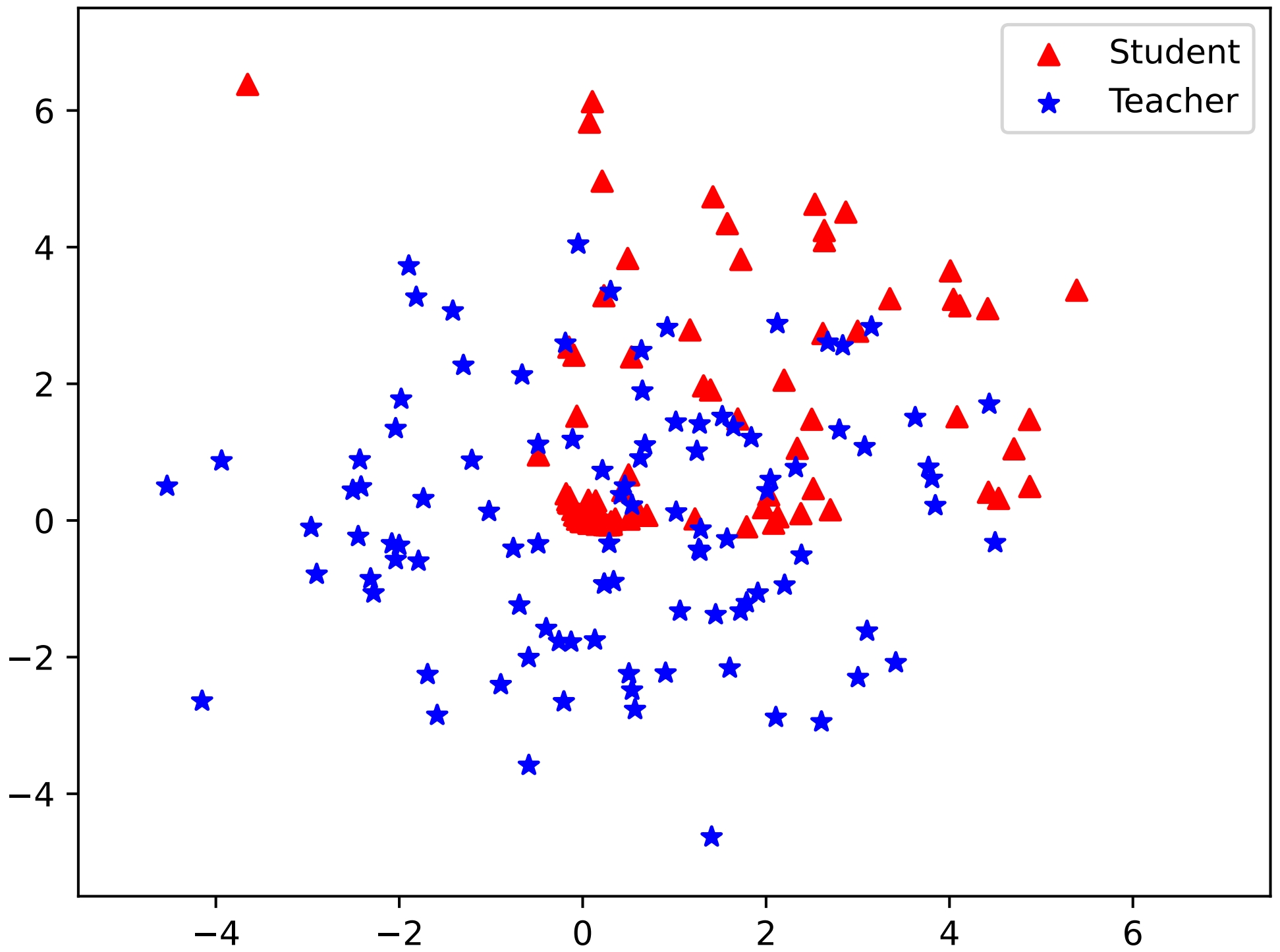}
		\end{minipage}
	}%
	%此处的空行很重要，想让图片在什么地方换行就在代码对应位置空行
	\subfigure[After KD (shared head)]{
		\begin{minipage}[t]{0.24\linewidth}
			\centering
			\includegraphics[width=\linewidth]{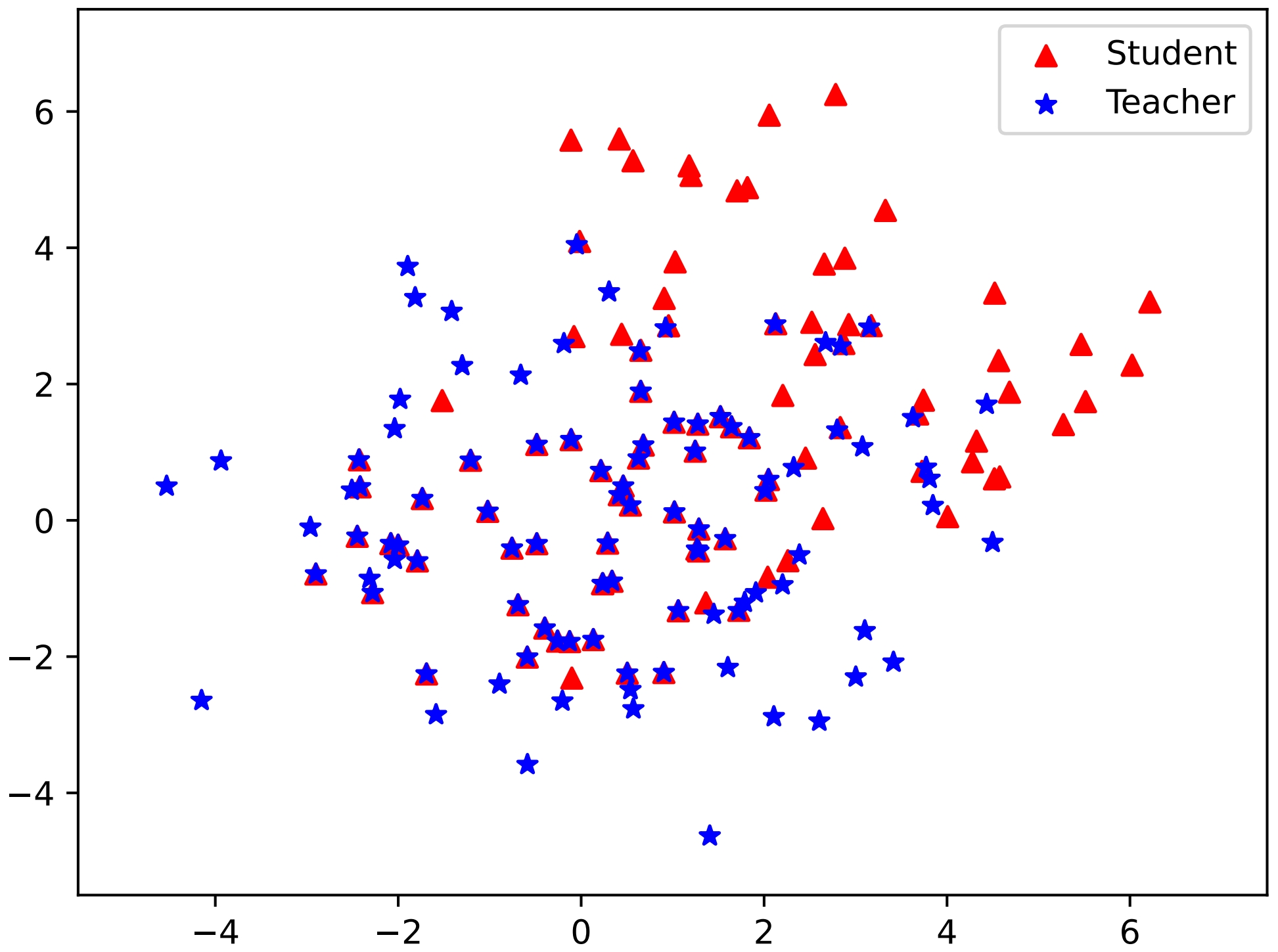}
		\end{minipage}
	}%
	\subfigure[Loss curves of KD]{
		\begin{minipage}[t]{0.24\linewidth}
			\centering
			\includegraphics[width=\linewidth]{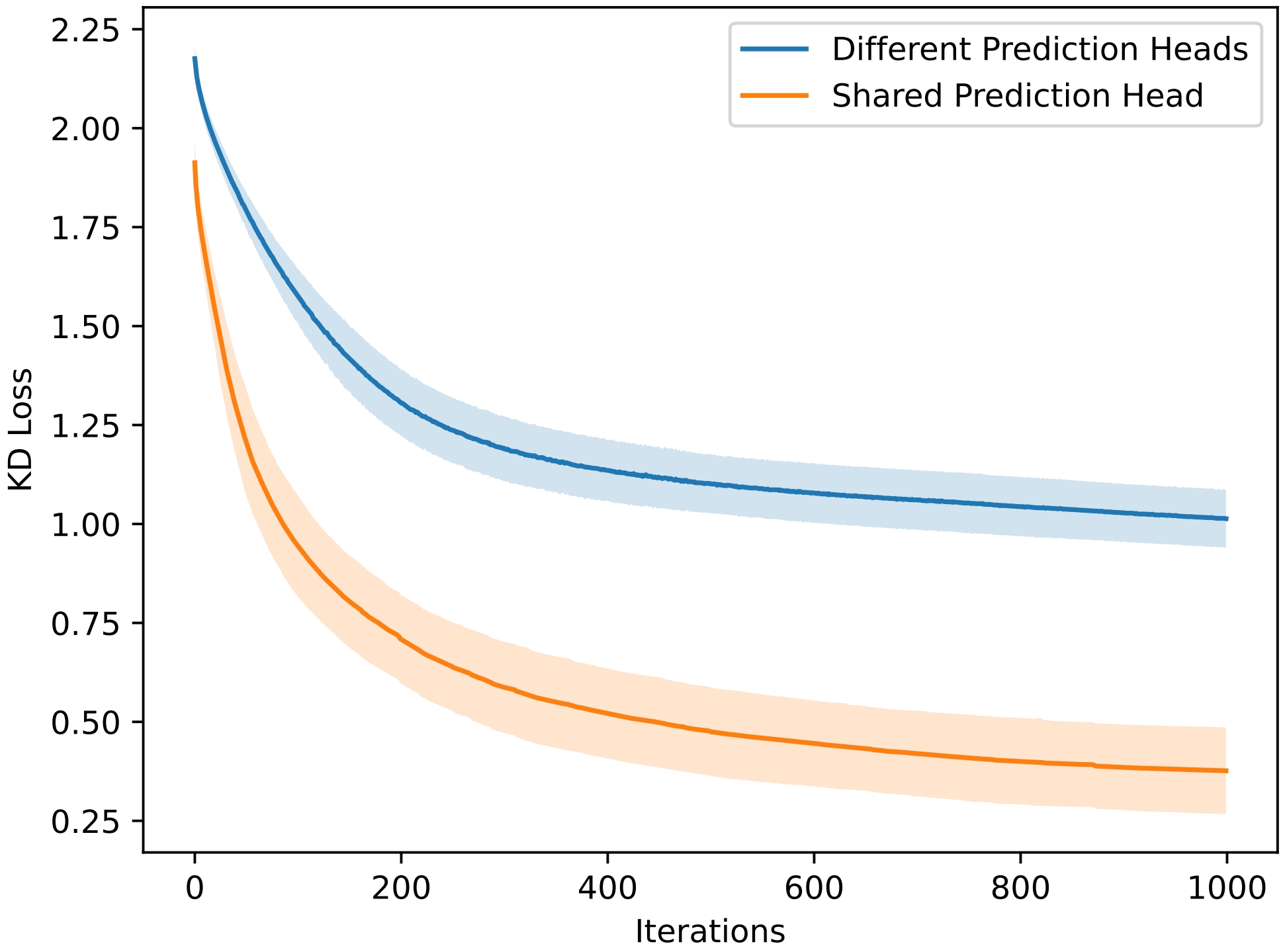}
		\end{minipage}
	}%
	\centering
	\caption{Simulation results with skewed reverse KL divergence as the KD objective. (a), (b) and (c) plot the {\color{red} student's hidden states} and the {\color{blue} teacher's hidden states} before and after the two KD processes. (d) shows the convergence curves of the KD objective in the two KD processes.}
	\label{fig:srkl_simulation}
\end{figure}

\begin{figure}[t]
	\centering
	\subfigure[Before KD]{
		\begin{minipage}[t]{0.24\linewidth}
			\centering
			\includegraphics[width=\linewidth]{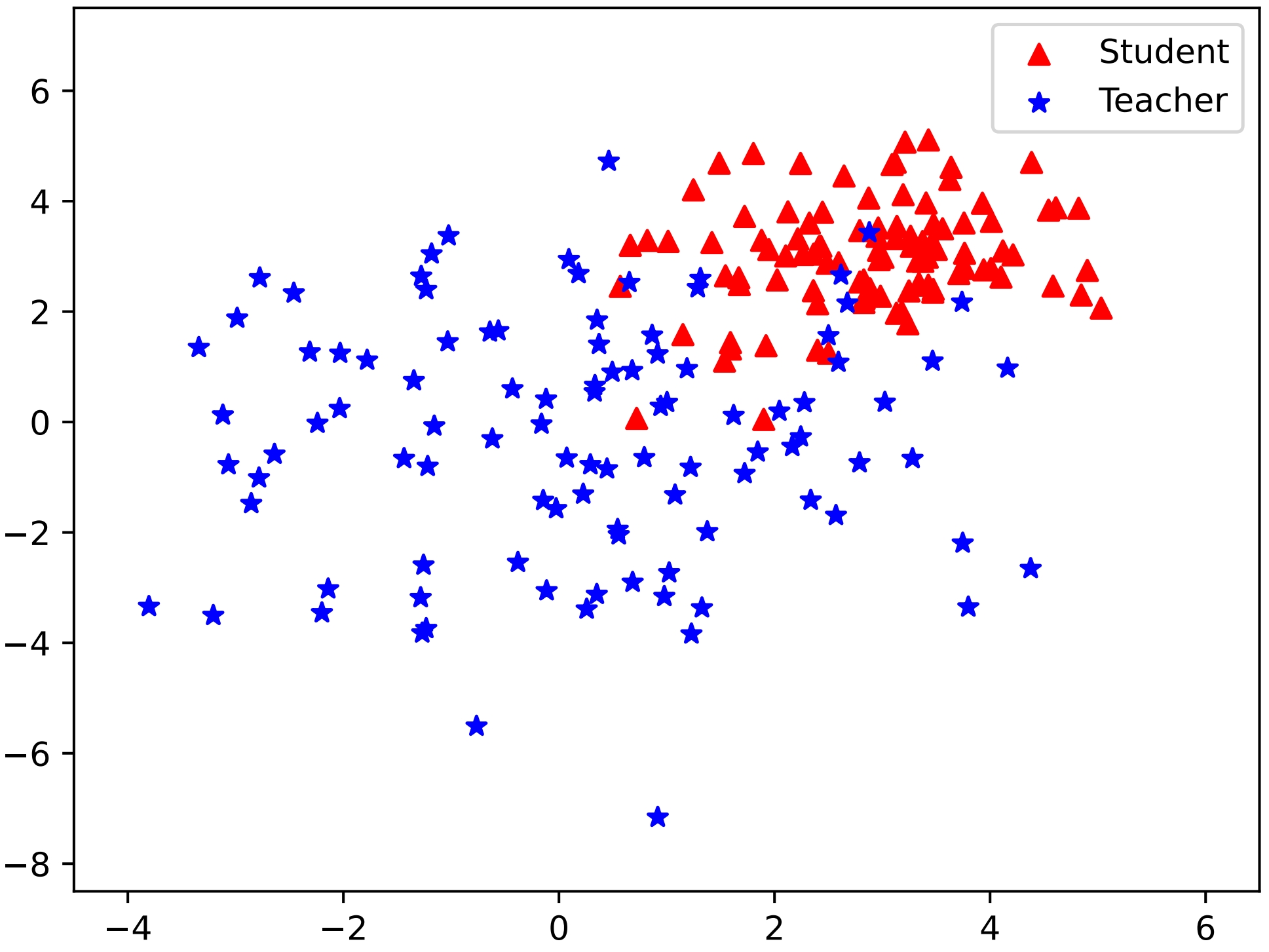}
		\end{minipage}
	}%
	\subfigure[After KD (different heads)]{
		\begin{minipage}[t]{0.24\linewidth}
			\centering
			\includegraphics[width=\linewidth]{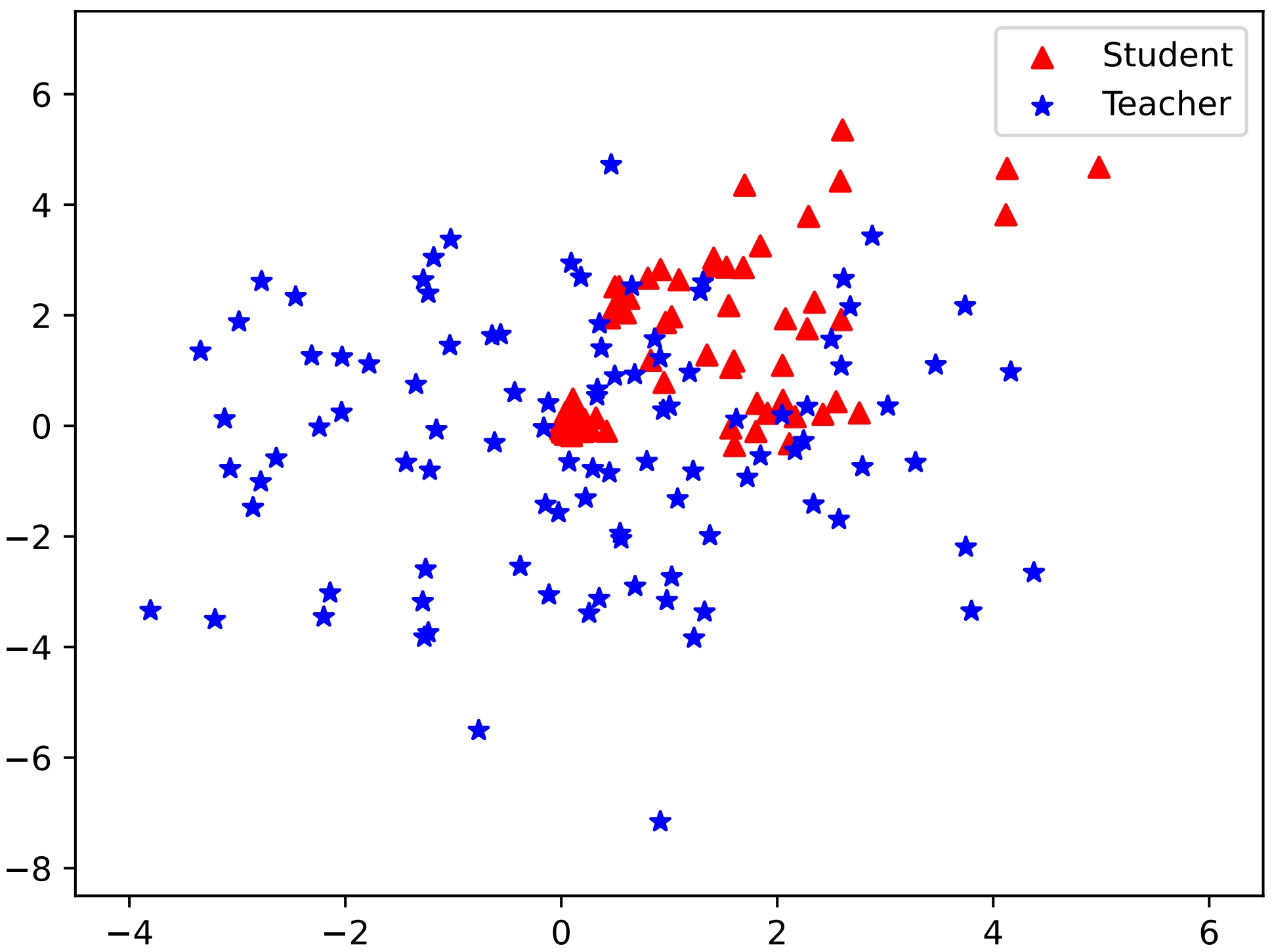}
		\end{minipage}
	}%
	%此处的空行很重要，想让图片在什么地方换行就在代码对应位置空行
	\subfigure[After KD (shared head)]{
		\begin{minipage}[t]{0.24\linewidth}
			\centering
			\includegraphics[width=\linewidth]{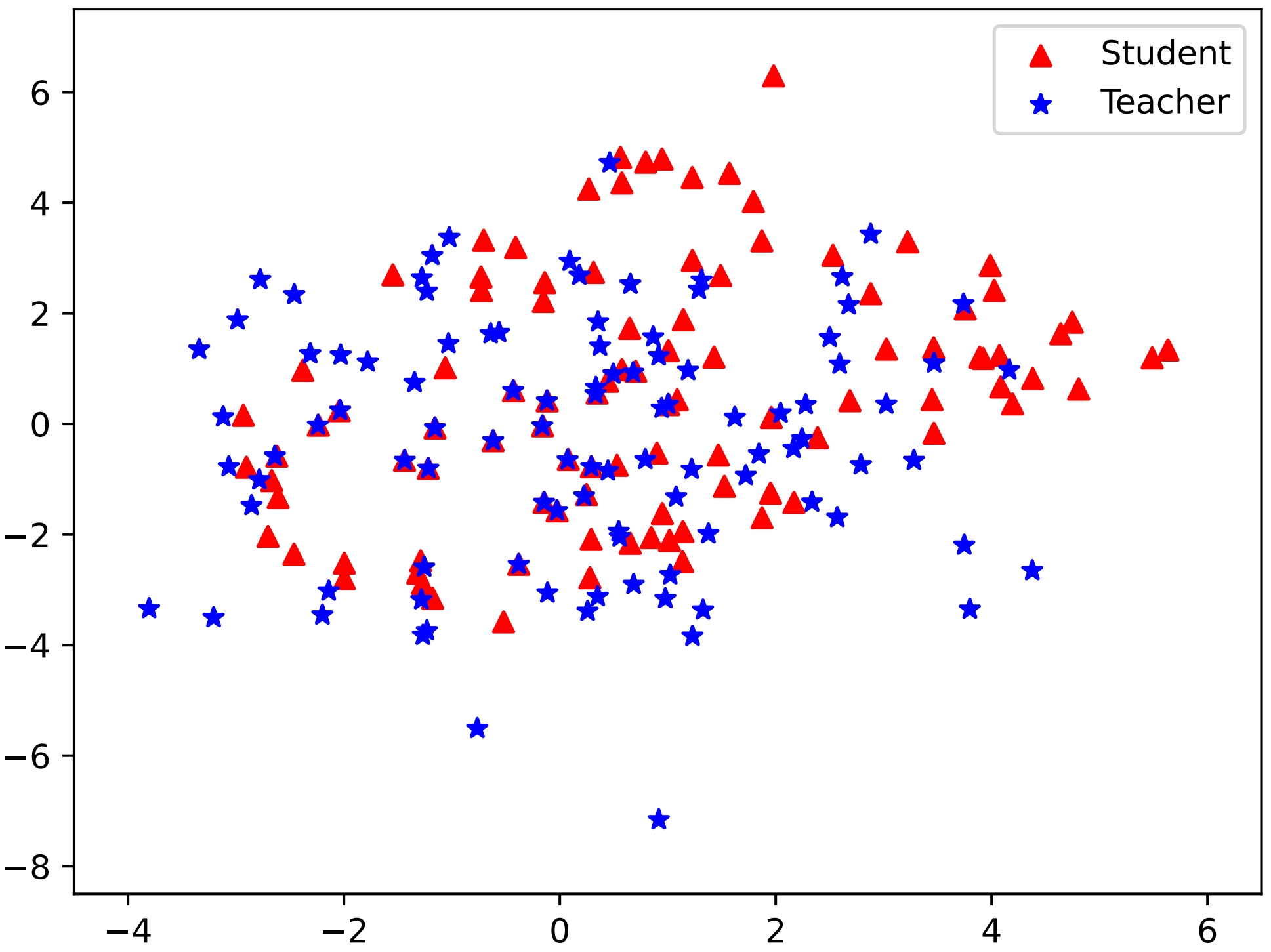}
		\end{minipage}
	}%
	\subfigure[Loss curves of KD]{
		\begin{minipage}[t]{0.24\linewidth}
			\centering
			\includegraphics[width=\linewidth]{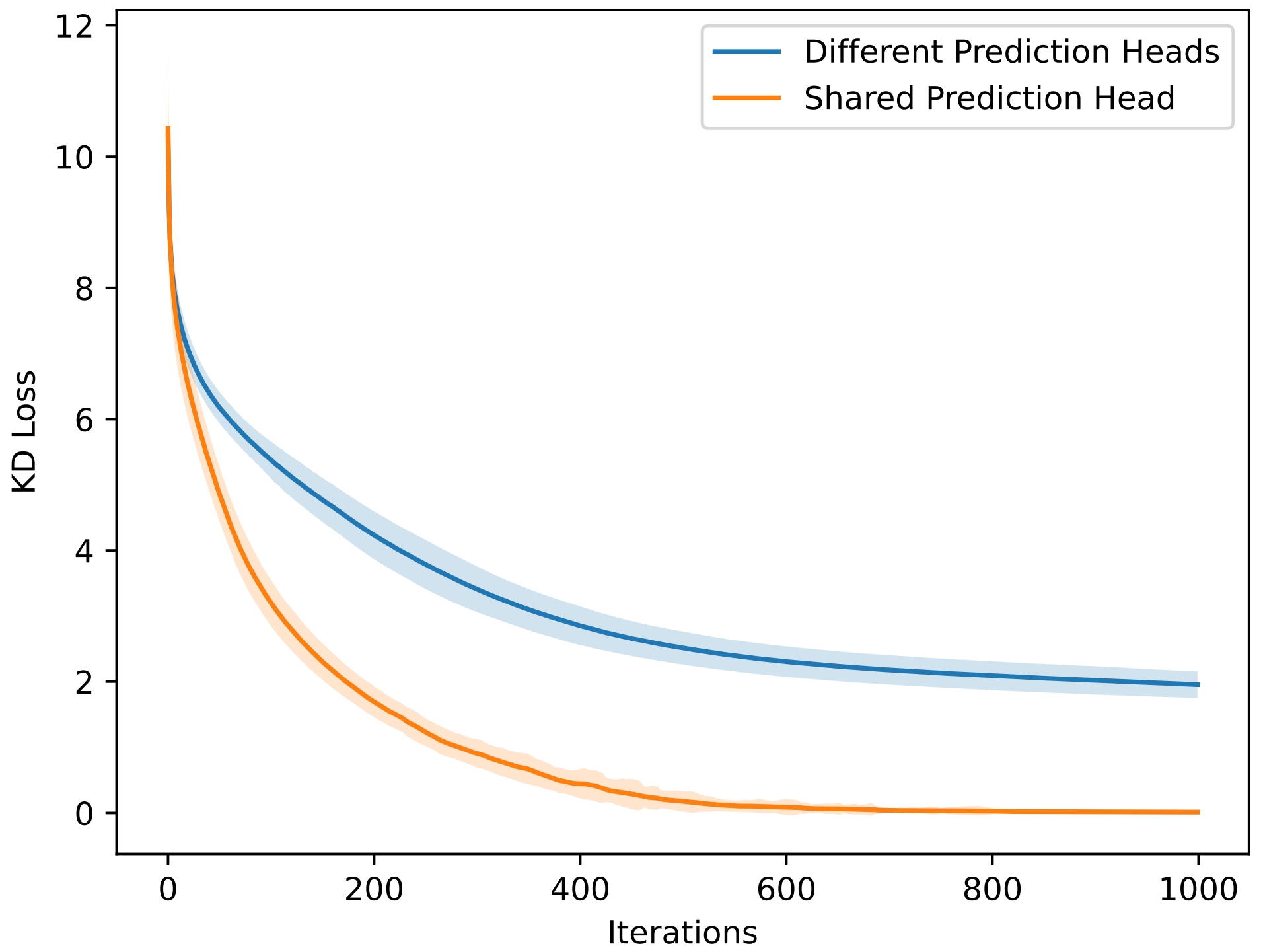}
		\end{minipage}
	}%
	\centering
	\caption{Simulation results with adaptive KL divergence as the KD objective. (a), (b) and (c) plot the {\color{red} student's hidden states} and the {\color{blue} teacher's hidden states} before and after the two KD processes. (d) shows the convergence curves of the KD objective in the two KD processes.}
	\label{fig:akl_simulation}
\end{figure}

We complement the remaining results of simulation experiments for the following objectives: reverse KL divergence, JS divergence, skewed KL divergence, skewed RKL divergence, and adaptive KL divergence.
The results are plotted in Figure \ref{fig:rkl_simulation}, Figure \ref{fig:js_simulation}, Figure \ref{fig:skl_simulation}, Figure \ref{fig:srkl_simulation} and Figure \ref{fig:akl_simulation}.
It is shown that no matter which distance function is used, the student after KD will have low representation similarity with the teacher and leave large margin to the minimum distance between the two distributions when using different prediction heads.
Thus, all these results lead to the consistent conclusion in \S\ref{sec:low_sim}, and also suggest that current KD framework may have inherent flaws on enhancing the similarity between the student model and the teacher model.
As a solution, unifying the output spaces by sharing the prediction head for teacher and student may achieve more effective KD process. 

\subsection{Pseudo Code for Simulation Experiments}
\label{sec:pseudo_code}
% \begin{minted}{c}
%     stu_logits = self.h1.matmul(self.e1.transpose(-1, -2))
%     tea_logits = self.h3.matmul(self.e3.transpose(-1, -2))
% \end{minted}
We also provide the pseudo code for re-implementing the key parts of our simulation experiments:

\lstset{
    numbers=none
}
\begin{lstlisting}
class Teacher(nn.Module):
    def __init__(self):
        super(Teacher, self).__init__()
        # the initial teacher hiddens are sampled from Gaussian Distribution N(0, 2)
        self.hidden = torch.randn(100, 2) * 2
        # the head contains 10000 classes
        self.head = torch.randn(10000, 2)  

class Student(nn.Module):
    def __init__(self):
        super(Student, self).__init__()
        # the initial student hiddens are sampled from Gaussian Distribution N(3, 1)
        self.hidden = nn.Parameter(torch.randn(100, 2) + 3)
        # the head contains 10000 classes
        self.head = nn.Parameter(torch.randn(10000, 2))  

def kd_with_different_head(student, teacher):
    student_logits = student.hidden.matmul(student.head.transpose(-1, -2))
    # calculating logits with the respective heads
    teacher_logits = teacher.hidden.matmul(teacher.head.transpose(-1, -2))  
    kd_loss = distance_func(student_logits, teacher_logits)
    return kd_loss

def kd_with_shared_head(student, teacher):
    student_logits = student.hidden.matmul(student.head.transpose(-1, -2))
    # calculating logits with the same head (student's head)
    teacher_logits = teacher.hidden.matmul(student.head.transpose(-1, -2))  
    kd_loss = distance_func(student_logits, teacher_logits)
    return kd_loss 
\end{lstlisting}
As shown in the code, we manually separate the hidden states of the student and teacher in initialization, so that the difference before and after KD will be more clear.
Besides, to unify the output spaces of the two models, we share the prediction head of the student with the teacher in ``\verb|kd_with_shared_head|''.
In this way, the output distributions of the student being optimized are as same as the ones in ``\verb|kd_with_different_head|'' and thus the results will be more comparable with the ones in ``\verb|kd_with_different_head|''.
The student models are optimized by the SGD optimizer with appropriate learning rates in $[1.0, 40.0]$ for different distance functions.

\section{Experimental Details} \label{sec:experiment_detail}
\subsection{Data}
All the test sets in our experiments are processed by \cite{gu23minillm}. 
For all these test sets, Dolly contains 500 samples, Self-Instruction \cite{wang23selfinst} contains 242 samples, Vicuna-Evaluation \cite{chiang23vicuna} contains 80 samples, Super-Natural Instructions \cite{wang22sni} contains 1694 samples with response lengths in $[11, +\infty]$, and Unnatural Instructions \cite{honovich23unni} contains 10000 samples with response lengths in $[11, +\infty]$.
\subsection{Training}
For GPT2-1.5B, we directly use the checkpoint released by \citet{gu23minillm}.
For other models, the detailed training configurations are listed in Table \ref{tab:train_config}.
Note that we do not use the pre-training corpus while distillation as \citep{gu23minillm} did for simplicity.
Each training requires several hours on 4$\times$RTX 3090 or 8$\times$RTX A4000.
\begin{table}[h]
    \centering
    \resizebox{0.7\linewidth}{!}{
        \begin{tabular}{c|cc|ccc}
            \bottomrule
            \multirow{2}{*}{\textbf{Settings}} & \multicolumn{2}{c|}{\textbf{KD for GPT2}} & \multicolumn{3}{c}{\textbf{KD for TinyLLaMA}} \\
            \cline{2-6}
            & GPT2 & Qwen1.5 & TinyLLaMA & LLaMA2 & Mistral \\ 
            \hline
            Epoch & 20 & 10 & 10 & 10 & 10 \\
            Learning Rate & 5e-4 & 2e-5 & 1e-3 & 1e-3 & 1e-3 \\
            Projector Learning Rate & 1e-3 & 1e-3 & 1e-3 & 1e-3 & 1e-3 \\
            Batch Size & 32 & 32 & 32 & 32 & 32 \\
            LR Scheduler & Cosine & Cosine & Cosine & Cosine & Cosine \\
            Fine-Tuning Method & Full & Full & LoRA & LoRA & LoRA \\
            Lora Rank & N/A & N/A & 256 & 256 & 256 \\
            Lora Alpha & N/A & N/A & 8 & 8 & 8 \\
            Lora Dropout & N/A & N/A & 0.1 & 0.1 & 0.1 \\
            \toprule
        \end{tabular}
    }
    \caption{Detailed training configurations of KD for GPT2 and TinyLLaMA.}
    \label{tab:train_config}
\end{table}

Besides, we combine the original cross-entropy loss on the target tokens in Eqn. \eqref{eq:ce_loss} and the KD loss in Eqn. \eqref{eq:kd_loss} and Eqn. \eqref{eq:dskd_loss} as the overall training loss for all the white-box KD methods in our main experiments:
\begin{equation}
    \mathcal{L} = 0.5 * \mathcal{L}_{ce} + 0.5 * \mathcal{L}_{(ds)kd}.
\end{equation}
% \begin{wrapfigure}[13]{r}{6.5cm}
%     \centering
%     \includegraphics[width=\linewidth]{temperature.png}
%     \caption{Rouge-L scores (\%) on the validation set for different temperature coefficients in KL divergence and reverse KL divergence.}
%     \label{fig:temperature}
% \end{wrapfigure}
\subsection{Evaluation}

For the evaluation, we use random sampling to decode the responses from all models.
For decoding, we set both the decoding temperature and top\_p to 1.0.  
Then, we generate the responses with random seeds in [10, 20, 30, 40, 50] and report the averaged Rouge-L scores of each seed following \citet{gu23minillm}.

\subsection{Effect of Temperature for KD} \label{sec:temperature}
\begin{figure}[h]
    \centering
    \includegraphics[width=0.4\linewidth]{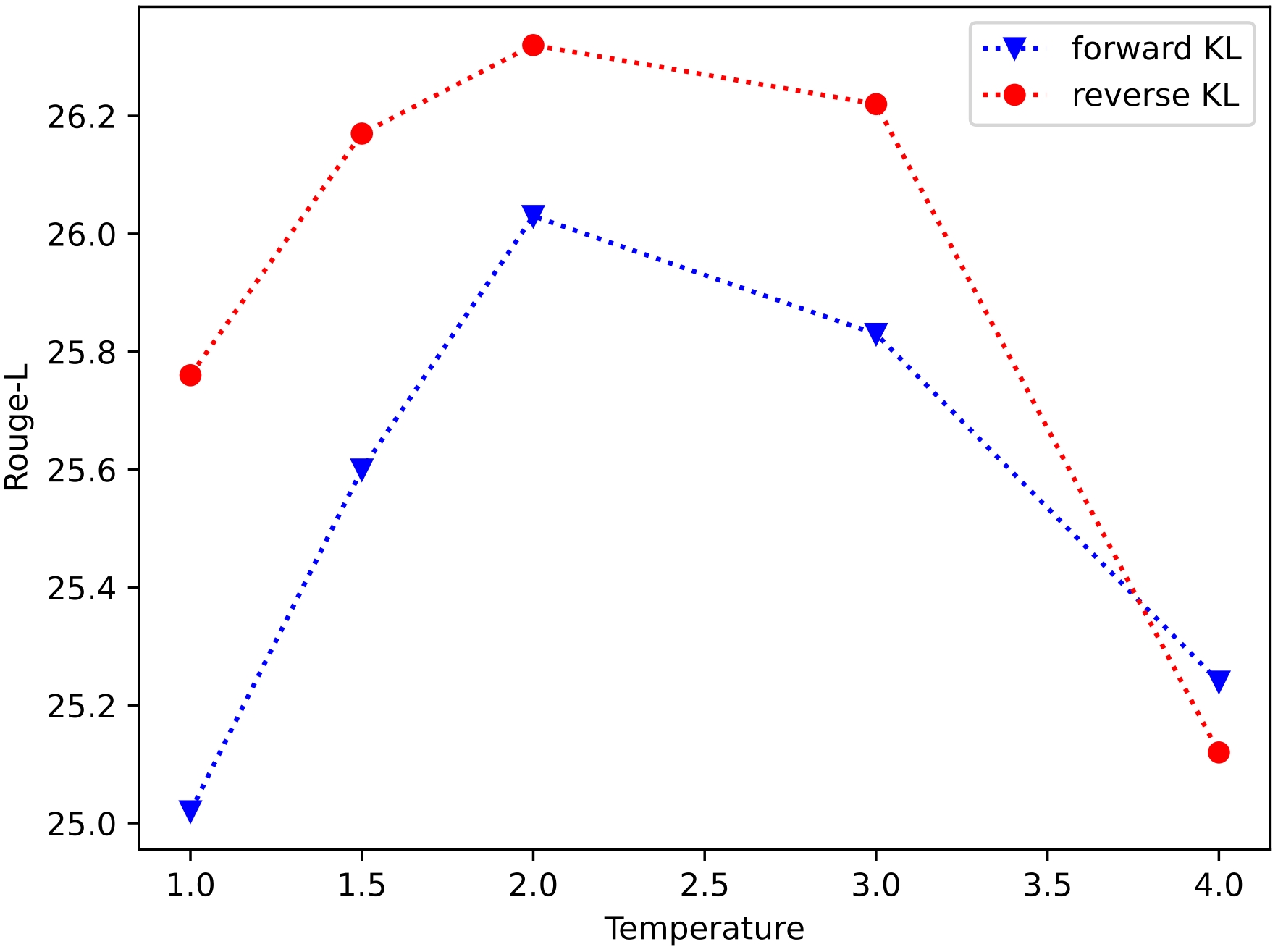}
    \caption{Rouge-L scores (\%) on the validation set for different temperature coefficients in KL divergence and reverse KL divergence.}
    \label{fig:temperature}
\end{figure}

As an important hyper-parameter in KD, the temperature coefficient $\tau$ significantly affects the final performance of KD.
As stated by the previous literature, a larger temperature (>1.0) will smooth the teacher's distribution and transfer more class relationship information to the student model. 
Thus, we search for the best temperatures among [1.0, 1.5, 2.0, 3.0, 4.0] for two representative objectives (\emph{i.e.}, KL divergence and reverse KL divergence) on the validation set and report the results in Figure \ref{fig:temperature}.
The results show that both objectives perform best when the temperature is 2.0.
Thus, we keep the temperature to 2.0 for all objectives in our experiments.

% \begin{wrapfigure}{r}{4cm}%靠文字内容的右侧
% \centering
% \includegraphics[width=0.15\textwidth]{WeChatlogo-6.jpg}
% \caption{\footnotesize 运动健康}
% \end{wrapfigure}

\section{Full Results}
We provide the full results of our main experiments in Table \ref{tab:full_results_gpt2} and Table \ref{tab:full_results_tinyllama}.
For KD between LLMs with the same vocabulary, we complement the detailed results of all distance functions in both the student and the teacher space.
For KD between LLMs with different vocabularies, we also present the full results of our DSKD with CMA for all the distance functions. 

As shown in Table \ref{tab:full_results_gpt2} and Table \ref{tab:full_results_tinyllama}, KD in the student space yields better performance than vanilla KD (in the different spaces) on all distance functions.
However, KD in the teacher space only leads to limited improvement for some distance functions.
The reason is that the student distribution $q_{\theta}^{s \rightarrow t}$ optimized by KD in the teacher space is different from the original student distribution $q_{\theta}$, and thus the KD process has no direct influence on $q_{\theta}$.
Nevertheless, we found that KL divergence has relatively good performance for KD in the teacher space.
Therefore, we directly choose KL divergence as the distance function for KD in the teacher space in our DSKD.

\label{sec:full_ablation}
\begin{table*}[h]
    \centering
    \resizebox{\linewidth}{!}{
        \begin{tabular}{lccccc|c}
            \bottomrule
            \textbf{Methods} & \textbf{Dolly} & \textbf{SelfInst} & \textbf{VicunaEval} & \textbf{S-NI} & \textbf{UnNI} & \textbf{Avg.} \\
            \hline
            \hline
            SFT & 22.94$_{\pm 0.28}$ & 10.11$_{\pm 0.36}$ & 15.17$_{\pm 0.63}$ & 16.21$_{\pm 0.19}$ & 18.68$_{\pm 0.09}$ & 16.62 \\
            \hline
            \multicolumn{7}{c}{\textbf{GPT2-1.5B $\rightarrow$ GPT2-120M (Same Vocabulary)}} \\
            \hline
            \rowcolor{lightgray} Teacher & {\color{midgray} 27.19$_{\pm 0.23}$} & {\color{midgray} 14.64$_{\pm 0.64}$} & {\color{midgray} 16.30$_{\pm 0.37}$} & {\color{midgray} 27.55$_{\pm 0.30}$} & {\color{midgray} 31.46$_{\pm 0.12}$} & {\color{midgray} 23.43} \\
            \hline
            SeqKD & 23.68$_{\pm 0.25}$ & 10.03$_{\pm 0.23}$ & 14.41$_{\pm 0.46}$ & 16.36$_{\pm 0.18}$ & 18.48$_{\pm 0.11}$ & 16.59 \\
            \hline
            KL & 24.54$_{\pm 0.48}$ & 10.43$_{\pm 0.24}$ & 15.66$_{\pm 0.42}$ & 17.24$_{\pm 0.27}$ & 20.28$_{\pm 0.18}$ & 17.63 \\
            % \rowcolor{lightgreen}
            \quad KL in Student Space & 23.83$_{\pm 0.30}$ & 10.46$_{\pm 0.36}$ & 15.79$_{\pm 0.51}$ & 18.82$_{\pm 0.31}$ & 21.08$_{\pm 0.07}$ & 18.00  \\
            \quad KL in Teacher Space & 24.07$_{\pm 0.67}$ & 10.34$_{\pm 0.38}$ & 14.94$_{\pm 0.24}$ & 18.83$_{\pm 0.25}$ & 21.02$_{\pm 0.11}$ & 17.84  \\
            \quad KL in Student Space + KL in Teacher Space & 24.70$_{\pm 0.24}$ & 10.65$_{\pm 0.30}$ & 15.67$_{\pm 0.30}$ & 19.51$_{\pm 0.21}$ & 22.94$_{\pm 0.07}$ & 18.69  \\
            \hline
            RKL & 24.38$_{\pm 0.55}$ & 10.73$_{\pm 0.61}$ & 15.71$_{\pm 0.39}$ & 17.31$_{\pm 0.11}$ & 20.96$_{\pm 0.12}$ & 17.82 \\
            \quad RKL in Student Space & 25.12$_{\pm 0.25}$ & 10.60$_{\pm 0.27}$ & 15.25$_{\pm 0.26}$ & 17.96$_{\pm 0.24}$ & 21.19$_{\pm 0.09}$ & 18.03  \\
            \quad RKL in Teacher Space & 23.54$_{\pm 0.33}$ & 10.48$_{\pm 0.55}$ & 15.21$_{\pm 0.52}$ & 16.59$_{\pm 0.18}$ & 19.49$_{\pm 0.16}$ & 17.06  \\
            % \rowcolor{lightgreen}
            \quad RKL in Student Space + KL in Teacher Space & 24.61$_{\pm 0.59}$ & 11.01$_{\pm 0.45}$ & 14.98$_{\pm 0.48}$ & 19.32$_{\pm 0.28}$ & 22.27$_{\pm 0.13}$ & 18.44  \\
            \hline
            JS & 23.86$_{\pm 0.14}$ & 10.20$_{\pm 0.40}$ & 15.50$_{\pm 0.23}$ & 16.20$_{\pm 0.23}$ & 19.17$_{\pm 0.06}$ & 16.98 \\
            \quad JS in Student Space & 24.46$_{\pm 0.34}$ & 10.02$_{\pm 0.24}$ & 15.59$_{\pm 0.46}$ & 16.53$_{\pm 0.19}$ & 19.25$_{\pm 0.14}$ & 17.17  \\
            \quad JS in Teacher Space & 23.28$_{\pm 0.52}$ & 9.76$_{\pm 0.37}$ & 15.08$_{\pm 0.26}$ & 15.89$_{\pm 0.20}$ & 18.34$_{\pm 0.12}$ & 16.47 \\
            % \rowcolor{lightgreen}
            \quad JS in Student Space + KL in Teacher Space & 24.61$_{\pm 0.27}$ & 11.41$_{\pm 0.35}$ & 15.40$_{\pm 0.28}$ & 18.94$_{\pm 0.20}$ & 21.48$_{\pm 0.17}$ & 18.37 \\
            \hline
            SKL \cite{ko24distillm} & 24.03$_{\pm 0.23}$ & 10.66$_{\pm 0.51}$ & 14.70$_{\pm 0.37}$ & 17.99$_{\pm 0.15}$ & 21.18$_{\pm 0.16}$ & 17.71 \\
            \quad SKL in Student Space & 24.06$_{\pm 0.38}$ & 11.03$_{\pm 0.18}$ & 15.11$_{\pm 0.44}$ & 18.67$_{\pm 0.27}$ & 21.13$_{\pm 0.05}$ & 18.00  \\
            \quad SKL in Teacher Space & 23.44$_{\pm 0.25}$ & 10.06$_{\pm 0.43}$ & 14.86$_{\pm 0.51}$ & 16.52$_{\pm 0.21}$ & 19.60$_{\pm 0.15}$ & 16.90 \\
            % \rowcolor{lightgreen}
            \quad SKL in Student Space + KL in Teacher Space & 25.24$_{\pm 0.28}$ & 10.50$_{\pm 0.13}$ & 15.76$_{\pm 0.43}$ & 18.34$_{\pm 0.44}$ & 20.87$_{\pm 0.11}$ & 18.14 \\
            \hline
            SRKL \cite{ko24distillm} & 24.48$_{\pm 0.19}$ & 10.35$_{\pm 0.38}$ & 14.88$_{\pm 0.24}$ & 16.53$_{\pm 0.23}$ & 19.68$_{\pm 0.05}$ & 17.19 \\
            \quad SRKL in Student Space & 24.84$_{\pm 0.08}$ & 10.50$_{\pm 0.59}$ & 15.16$_{\pm 0.30}$ & 16.80$_{\pm 0.26}$ & 20.04$_{\pm 0.05}$ & 17.47  \\
            \quad SRKL in Teacher Space & 23.10$_{\pm 0.39}$ & 10.00$_{\pm 0.42}$ & 14.83$_{\pm 0.39}$ & 16.07$_{\pm 0.34}$ & 18.45$_{\pm 0.17}$ & 16.49  \\
            % \rowcolor{lightgreen}
            \quad SRKL in Student Space + KL in Teacher Space & 25.23$_{\pm 0.25}$ & 11.19$_{\pm 0.22}$ & 15.91$_{\pm 0.45}$ & 17.92$_{\pm 0.16}$ & 21.20$_{\pm 0.12}$ & 18.29 \\
            \hline
            AKL \cite{wu2024rethinking} & 24.75$_{\pm 0.60}$ & 10.46$_{\pm 0.24}$ & 15.37$_{\pm 0.41}$ & 17.48$_{\pm 0.17}$ & 20.11$_{\pm 0.05}$ & 17.63 \\
            \quad AKL in Student Space & 25.08$_{\pm 0.36}$ & 10.70$_{\pm 0.15}$ & 14.56$_{\pm 0.74}$ & 17.80$_{\pm 0.20}$ & 20.72$_{\pm 0.11}$ & 17.77  \\
            \quad AKL in Teacher Space & 23.82$_{\pm 0.60}$ & 10.10$_{\pm 0.59}$ & 15.40$_{\pm 0.16}$ & 17.04$_{\pm 0.16}$ & 20.13$_{\pm 0.09}$ & 17.30  \\
            % DSKD (ours) & & & & & \\
            % \rowcolor{lightgreen}
            \quad AKL in Student Space + KL in Teacher Space & 25.13$_{\pm 0.14}$ & 10.63$_{\pm 0.43}$ & 16.18$_{\pm 0.35}$ & 18.58$_{\pm 0.48}$ & 21.45$_{\pm 0.16}$ & 18.39 \\
            \hline
            \multicolumn{7}{c}{\textbf{Qwen1.5-1.8B $\rightarrow$ GPT2-120M (Different Vocabulary)}} \\
            \hline
            \rowcolor{lightgray} Teacher & {\color{midgray} 27.19$_{\pm 0.23}$} & {\color{midgray} 14.64$_{\pm 0.64}$} & {\color{midgray} 16.30$_{\pm 0.37}$} & {\color{midgray} 27.55$_{\pm 0.30}$} & {\color{midgray} 31.42$_{\pm 0.11}$} & {\color{midgray} 23.42} \\
            SeqKD & 23.40$_{\pm 0.21}$ & 9.36$_{\pm 0.38}$ & 15.37$_{\pm 0.35}$ & 15.16$_{\pm 0.17}$ & 17.34$_{\pm 0.11}$ & 16.13  \\
            MinED \cite{wan24fusellm} & 24.41$_{\pm 0.61}$ & 10.60$_{\pm 0.39}$ & 15.86$_{\pm 0.42}$ & 16.76$_{\pm 0.28}$ & 19.68$_{\pm 0.12}$ & 17.46 \\
            ULD \cite{boizard2024uld} & 23.77$_{\pm 0.41}$ & 9.67$_{\pm 0.50}$ & 14.99$_{\pm 0.55}$ & 17.60$_{\pm 0.21}$ & 19.49$_{\pm 0.12}$ & 17.11  \\
            % DSKD + Align (ours) & & & & & \\
            \hline
            DSKD-CMA-KL (ours) & 24.73$_{\pm 0.47}$ & 11.15$_{\pm 0.34}$ & 15.31$_{\pm 0.38}$ & 17.20$_{\pm 0.24}$ & 20.57$_{\pm 0.08}$ & 17.79  \\
            DSKD-CMA-RKL (ours) & 23.99$_{\pm 0.29}$ & 10.89$_{\pm 0.46}$ & 15.15$_{\pm 0.28}$ & 17.82$_{\pm 0.11}$ & 21.05$_{\pm 0.13}$ & 17.78  \\
            DSKD-CMA-JS (ours) & 23.95$_{\pm 0.29}$ & 10.44$_{\pm 0.60}$ & 15.38$_{\pm 0.23}$ & 16.69$_{\pm 0.14}$ & 20.27$_{\pm 0.10}$ & 17.35  \\
            DSKD-CMA-SKL (ours) & 24.67$_{\pm 0.13}$ & 10.82$_{\pm 0.46}$ & 15.30$_{\pm 0.51}$ & 17.95$_{\pm 0.28}$ & 20.65$_{\pm 0.13}$ & 17.88  \\
            DSKD-CMA-SRKL (ours) & 25.23$_{\pm 0.17}$ & 10.99$_{\pm 0.26}$ & 15.56$_{\pm 0.41}$ & 17.76$_{\pm 0.23}$ & 20.54$_{\pm 0.07}$ & 18.02 \\
            DSKD-CMA-AKL (ours) & 24.72$_{\pm 0.33}$ & 10.67$_{\pm 0.29}$ & 15.84$_{\pm 0.67}$ & 16.59$_{\pm 0.25}$ & 19.78$_{\pm 0.10}$ & 17.52 \\
            \toprule
        \end{tabular}
    }
    \caption{Detailed Rouge-L scores (\%) of all our models on several benchmarks with GPT2-120M as the student. We present the mean values and the standard deviations among 5 random seeds. The average scores (\textbf{Avg.}) on all benchmarks are also listed. ``XX in Student Space + KL in Teacher Space'' represents our DSKD with XX as the distance function in Eqn. \eqref{eq:stuside_kd_loss}.}
    \label{tab:full_results_gpt2}
\end{table*}

\section{Details and Full Results for GPT-4 Evaluation}
We use the API of \texttt{gpt4-turbo-0409} to evaluate the quality of the responses.
As we conduct pairwise comparison between the responses from two models, to alleviate the order bias in the evaluation process of GPT-4, we randomly shuffle the two responses as the Response A/B in the system prompts.
\label{sec:llm_eval_full}
\begin{figure}[h]
    \centering
    \includegraphics[width=0.8\linewidth]{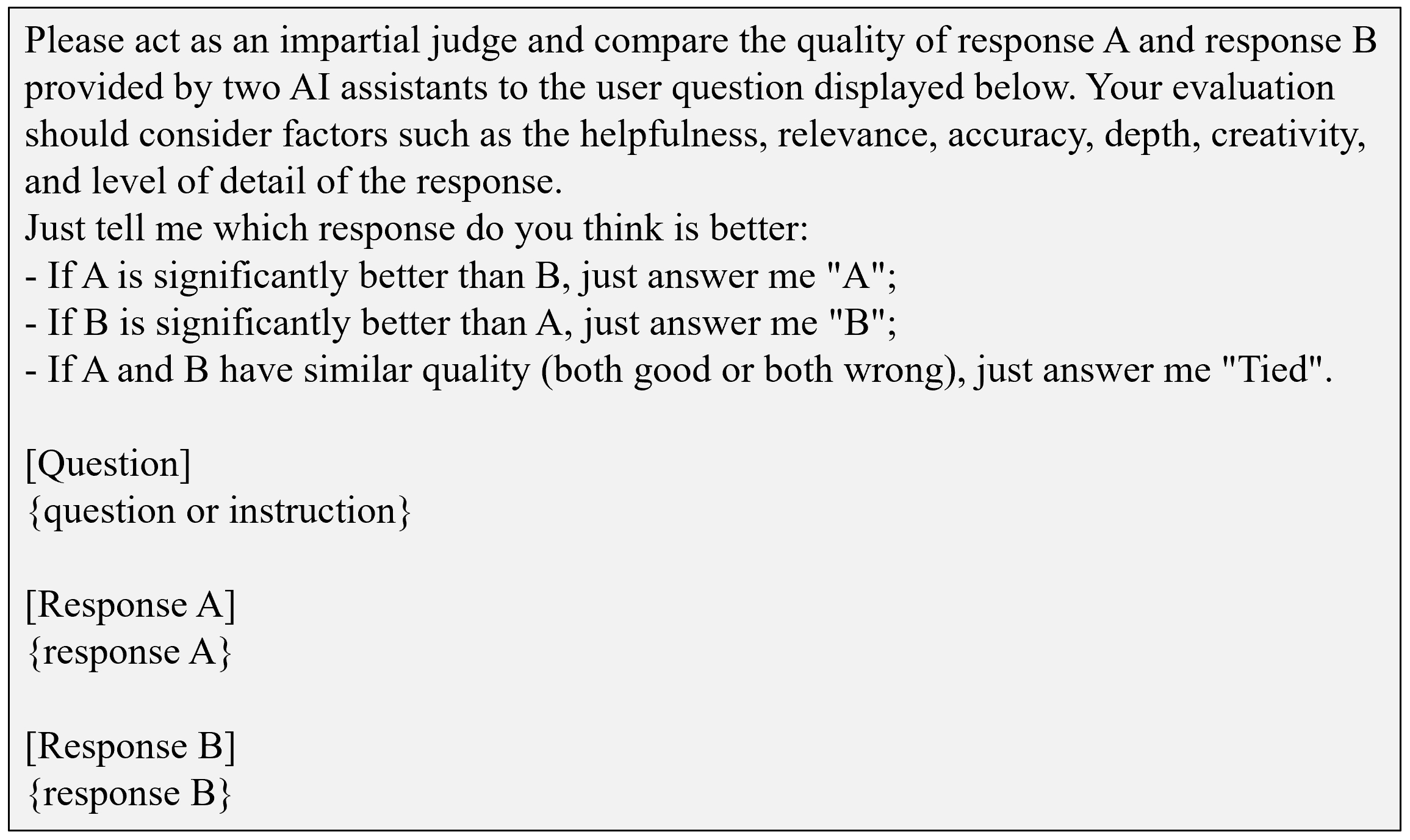}
    \caption{Prompt for GPT-4 Evaluation.}
    \label{fig:prompt}
\end{figure}

\begin{figure}[h]
    \centering
    \includegraphics[width=0.6\linewidth]{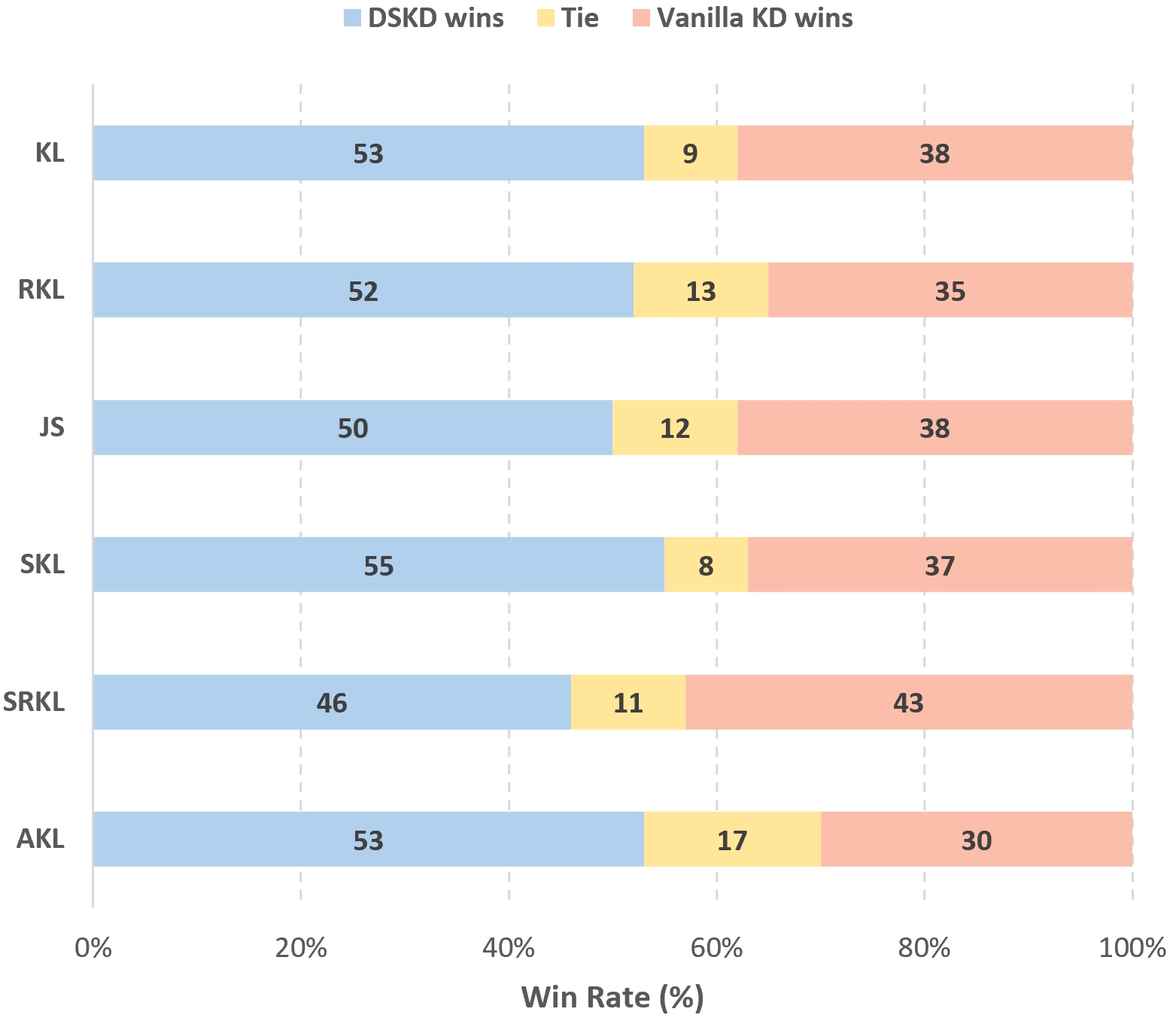}
    \caption{GPT-4 Evaluation Results for all the distance functions.}
    \label{fig:llm_eval_all}
\end{figure}

The full results for GPT-4 Evaluation on all distance functions are shown in Figure \ref{fig:llm_eval_all}.
For all distance functions, the students trained by our DSKD always win more than the student trained by the current white-box KD framework, indicating the consistent superiority of our DSKD framework on existing distance functions.

\section{Details of the Distance between Representation Structure}
\label{sec:repr_distance}
Since the student models and the teacher models generally have different dimensions on representations, it is difficult to directly measure the representation similarity between the student and the teacher.
Thus, we calculate the similarity on the structure of sentences in their own representation spaces of the student and the teacher.
Specifically, given a sentence with $n$ tokens, we calculate structure matrices with both the cosine similarity and normalized inner-product values between the output hidden states of this sentence:
\begin{equation} \label{eq:cosine}
    \mathcal{M}_{cosine}(i, j)=\frac{{h_i}^\top h_j}{|h_i| |h_j|} \in \mathbb{R}^{n \times n},
\end{equation}
\begin{equation} \label{eq:inner_prod}
    \mathcal{M}_{prod}(i, j)=\frac{{h_i}^{\top} h_j}{\sum_k {h_i}^{\top} h_k} \in \mathbb{R}^{n \times n},
\end{equation}
where $\mathcal{M}_{cosine}$ and $\mathcal{M}_{prod}$ are structure matrices calculated by cosine and normalized inner-product between output hidden states, respectively.
Then we calculate the L1 distance between the matrices of the student and the teacher:
\begin{equation} \label{eq:d_cosine}
    \mathcal{D}_{cosine}=\sum_i^n\sum_j^n |\mathcal{M}^t_{cosine}(i, j) - \mathcal{M}^s_{cosine}(i, j)|,
\end{equation}
\begin{equation} \label{eq:d_inner_prod}
    \mathcal{D}_{prod}=\sum_i^n\sum_j^n |\mathcal{M}^t_{prod}(i, j) - \mathcal{M}^s_{prod}(i, j)|.
\end{equation}
The smaller distance values means the representations of the student and the teacher are more similar.
In Figure \ref{fig:repr_sim}, we calculate and average the two distances $\mathcal{D}_{cosine}$ and $\mathcal{D}_{prod}$ on 1000 samples in the training set for GPT2 models that trained without KD (SFT), trained by the current white-box KD framework (Vanilla KD) and trained by our DSKD framework (DSKD).

\begin{table*}[h]
    \centering
    \resizebox{\linewidth}{!}{
        \begin{tabular}{lccccc|c}
            \bottomrule
            \textbf{Methods} & \textbf{Dolly} & \textbf{SelfInst} & \textbf{VicunaEval} & \textbf{S-NI} & \textbf{UnNI} & \textbf{Avg.} \\
            \hline
            \hline
            SFT & 23.20$_{\pm 0.13}$ & 14.88$_{\pm 0.54}$ & 16.42$_{\pm 0.35}$ & 27.79$_{\pm 0.27}$ & 26.12$_{\pm 0.11}$ & 21.68 \\
            \hline
            \multicolumn{7}{c}{\textbf{LLaMA2-7B $\rightarrow$ TinyLLaMA-1.1B (Same Vocabulary)}} \\
            \hline
            \rowcolor{lightgray} Teacher & 28.32$_{\pm 0.46}$ & 20.95$_{\pm 0.69}$ & 18.76$_{\pm 0.35}$ & 32.05$_{\pm 0.28}$ & 32.41$_{\pm 0.12}$ & 26.50 \\
            \hline
            SeqKD & 23.21$_{\pm 0.22}$ & 16.46$_{\pm 0.72}$ & 16.58$_{\pm 0.38}$ & 26.33$_{\pm 0.26}$ & 27.69$_{\pm 0.10}$ & 22.05 \\
            \hline
            KL & 25.46$_{\pm 0.63}$ & 17.21$_{\pm 0.25}$ & 16.43$_{\pm 0.53}$ & 29.27$_{\pm 0.29}$ & 29.28$_{\pm 0.09}$ & 23.53 \\
            % \rowcolor{lightgreen}
            \quad KL in Student Space & 26.20$_{\pm 0.30}$ & 18.69$_{\pm 0.72}$ & 17.71$_{\pm 0.43}$ & 32.40$_{\pm 0.21}$ & 29.94$_{\pm 0.09}$ & 24.99  \\
            \quad KL in Teacher Space & 22.86$_{\pm 0.77}$ & 15.80$_{\pm 0.53}$ & 15.90$_{\pm 0.22}$ & 27.58$_{\pm 0.29}$ & 28.03$_{\pm 0.20}$ & 22.04  \\
            \quad KL in Student Space + KL in Teacher Space & 26.31$_{\pm 0.26}$ & 18.27$_{\pm 0.56}$ & 18.04$_{\pm 0.37}$ & 31.43$_{\pm 0.26}$ & 31.20$_{\pm 0.09}$ & 25.05 \\
            \hline
            RKL & 24.49$_{\pm 0.41}$ & 17.14$_{\pm 0.61}$ & 16.87$_{\pm 0.26}$ & 29.50$_{\pm 0.28}$ & 29.36$_{\pm 0.08}$ & 23.47 \\
            \quad RKL in Student Space & 26.74$_{\pm 0.36}$ & 19.16$_{\pm 0.29}$ & 18.85$_{\pm 0.41}$ & 31.76$_{\pm 0.42}$ & 31.01$_{\pm 0.06}$ & 25.50  \\
            \quad RKL in Teacher Space & 22.60$_{\pm 0.43}$ & 16.04$_{\pm 1.15}$ & 15.81$_{\pm 0.40}$ & 28.88$_{\pm 0.23}$ & 28.86$_{\pm 0.10}$ & 22.44  \\
            \quad RKL in Student Space + KL in Teacher Space & 26.93$_{\pm 0.34}$ & 18.14$_{\pm 0.54}$ & 18.81$_{\pm 0.39}$ & 31.79$_{\pm 0.31}$ & 32.49$_{\pm 0.11}$ & 25.63 \\
            \hline
            JS & 24.03$_{\pm 0.31}$ & 15.75$_{\pm 0.51}$ & 16.64$_{\pm 0.30}$ & 28.08$_{\pm 0.10}$ & 28.68$_{\pm 0.08}$ & 22.62 \\
            % \rowcolor{lightgreen}
            \quad JS in Student Space & 23.86$_{\pm 0.26}$ & 17.16$_{\pm 0.85}$ & 16.98$_{\pm 0.39}$ & 27.61$_{\pm 0.27}$ & 27.65$_{\pm 0.08}$ & 22.64  \\
            \quad JS in Teacher Space & 22.74$_{\pm 0.34}$ & 15.28$_{\pm 0.74}$ & 16.33$_{\pm 0.26}$ & 26.54$_{\pm 0.28}$ & 26.07$_{\pm 0.14}$ & 21.39  \\
            \quad JS in Student Space + KL in Teacher Space & 24.79$_{\pm 0.42}$ & 17.10$_{\pm 0.47}$ & 16.78$_{\pm 0.20}$ & 29.06$_{\pm 0.18}$ & 29.47$_{\pm 0.22}$ & 23.44 \\
            \hline
            SKL \cite{ko24distillm} & 24.14$_{\pm 0.53}$ & 15.98$_{\pm 0.72}$ & 16.89$_{\pm 0.22}$ & 29.30$_{\pm 0.18}$ & 28.71$_{\pm 0.12}$ & 23.01 \\
            \quad SKL in Student Space & 25.15$_{\pm 0.24}$ & 17.16$_{\pm 0.84}$ & 17.27$_{\pm 0.18}$ & 29.19$_{\pm 0.19}$ & 28.98$_{\pm 0.20}$ & 23.55  \\
            \quad SKL in Teacher Space & 22.72$_{\pm 0.75}$ & 15.88$_{\pm 0.64}$ & 15.89$_{\pm 0.41}$ & 28.37$_{\pm 0.23}$ & 26.84$_{\pm 0.15}$ & 21.94  \\
            \quad SKL in Student Space + KL in Teacher Space & 25.88$_{\pm 0.22}$ & 17.59$_{\pm 0.56}$ & 17.17$_{\pm 0.34}$ & 29.52$_{\pm 0.33}$ & 30.69$_{\pm 0.16}$ & 24.17 \\
            \hline
            SRKL \cite{ko24distillm} & 24.28$_{\pm 0.58}$ & 16.91$_{\pm 0.67}$ & 16.88$_{\pm 0.20}$ & 29.55$_{\pm 0.19}$ & 28.64$_{\pm 0.21}$ & 23.25 \\
            \quad SRKL in Student Space & 25.92$_{\pm 0.39}$ & 16.76$_{\pm 0.71}$ & 17.13$_{\pm 0.46}$ & 29.69$_{\pm 0.17}$ & 28.67$_{\pm 0.04}$ & 23.64  \\
            \quad SRKL in Teacher Space & 22.88$_{\pm 0.57}$ & 16.40$_{\pm 0.46}$ & 16.24$_{\pm 0.40}$ & 27.23$_{\pm 0.37}$ & 27.16$_{\pm 0.04}$ & 21.98  \\
            \quad SRKL in Student Space + KL in Teacher Space & 25.44$_{\pm 0.22}$ & 17.34$_{\pm 0.69}$ & 17.19$_{\pm 0.34}$ & 30.29$_{\pm 0.29}$ & 31.23$_{\pm 0.13}$ & 24.30 \\
            \hline
            AKL \cite{wu2024rethinking} & 24.80$_{\pm 0.70}$ & 16.79$_{\pm 1.09}$ & 16.80$_{\pm 0.44}$ & 29.29$_{\pm 0.35}$ & 28.81$_{\pm 0.09}$ & 23.30 \\
            \quad AKL in Student Space & 26.07$_{\pm 0.51}$ & 19.57$_{\pm 0.83}$ & 17.57$_{\pm 0.46}$ & 34.50$_{\pm 0.33}$ & 33.45$_{\pm 0.15}$ & 26.23  \\
            \quad AKL in Teacher Space & 22.81$_{\pm 0.56}$ & 16.33$_{\pm 0.73}$ & 16.00$_{\pm 0.14}$ & 27.05$_{\pm 0.15}$ & 28.09$_{\pm 0.19}$ & 22.05  \\
            % DSKD (ours) & & & & & \\
            % \rowcolor{lightlightgray}
            \quad AKL in Student Space + KL in Teacher Space & 26.33$_{\pm 0.45}$ & 20.17$_{\pm 0.46}$ & 17.43$_{\pm 0.48}$ & 34.93$_{\pm 0.39}$ & 34.40$_{\pm 0.20}$ & 26.65 \\
            \hline
            % \rowcolor{lightpink}
            % \rowcolor{lightgray}
            \multicolumn{7}{c}{\textbf{Mistral-7B $\rightarrow$ TinyLLaMA-1.1B (Different Vocabularies)}} \\
            \hline
            \rowcolor{lightgray}
            Teacher & 31.56$_{\pm 0.19}$ & 25.10$_{\pm 0.36}$ & 20.50$_{\pm 0.32}$ & 36.07$_{\pm 0.24}$ & 36.27$_{\pm 0.15}$ & 29.90 \\
            \hline
            SeqKD & 23.56$_{\pm 0.39}$ & 15.87$_{\pm 0.54}$ & 15.99$_{\pm 0.55}$ & 25.50$_{\pm 0.37}$ & 26.64$_{\pm 0.09}$ & 21.51 \\
            MinED \cite{wan24fusellm} & 20.96$_{\pm 0.51}$ & 14.49$_{\pm 0.35}$ & 15.98$_{\pm 0.45}$ & 27.21$_{\pm 0.13}$ & 26.47$_{\pm 0.11}$ & 21.77 \\
            ULD \cite{boizard2024uld} \qquad & 22.80$_{\pm 0.28}$ & 15.93$_{\pm 0.74}$ & 16.43$_{\pm 0.60}$ & 26.94$_{\pm 0.28}$ & 24.83$_{\pm 0.13}$ & 20.64 \\
            % DSKD + Align (ours) & & & & & \\
            \hline
            DSKD-CMA-KL (ours) & 26.52$_{\pm 0.45}$ & 17.90$_{\pm 0.69}$ & 18.20$_{\pm 0.59}$ & 30.66$_{\pm 0.39}$ & 31.03$_{\pm 0.11}$ & 24.86 \\
            DSKD-CMA-RKL (ours) & 25.41$_{\pm 0.18}$ & 18.31$_{\pm 0.45}$ & 16.83$_{\pm 0.46}$ & 34.79$_{\pm 0.16}$ & 34.05$_{\pm 0.12}$ & 25.88 \\
            DSKD-CMA-JS (ours) & 24.09$_{\pm 0.71}$ & 16.77$_{\pm 0.75}$ & 16.96$_{\pm 0.27}$ & 30.01$_{\pm 0.15}$ & 30.00$_{\pm 0.10}$ & 23.56 \\
            DSKD-CMA-SKL (ours) & 25.28$_{\pm 0.24}$ & 17.33$_{\pm 0.62}$ & 17.57$_{\pm 0.43}$ & 30.27$_{\pm 0.30}$ & 31.14$_{\pm 0.35}$ & 24.32 \\
            DSKD-CMA-SRKL (ours) & 24.87$_{\pm 0.50}$ & 17.63$_{\pm 0.53}$ & 17.16$_{\pm 0.24}$ & 29.77$_{\pm 0.19}$ & 30.78$_{\pm 0.14}$ & 24.04 \\
            DSKD-CMA-AKL (ours) & 26.45$_{\pm 0.56}$ & 19.57$_{\pm 0.69}$ & 17.95$_{\pm 0.55}$ & 35.99$_{\pm 0.19}$ & 35.00$_{\pm 0.16}$ & 26.99 \\
            \toprule
        \end{tabular}
    }
    \caption{Rouge-L scores (\%) of all models on several benchmarks with TinyLLaMA-1.1B as the student. We present the mean values and the standard deviations among 5 random seeds. The average scores (\textbf{Avg.}) on all benchmarks are also listed. ``XX in Student Space + KL in Teacher Space'' represents our DSKD with XX as the distance function in Eqn. \eqref{eq:stuside_kd_loss}.}
    \label{tab:full_results_tinyllama}
\end{table*}

\end{document}